\documentclass{article}

\usepackage{microtype}
\usepackage{graphicx}
\usepackage{subfigure}
\usepackage{booktabs} % for professional tables
\usepackage{lipsum}
\usepackage{enumitem}
\usepackage{hyperref}

\usepackage[accepted]{icml2025}

\newcommand{\methodname}{AQUA-KV}
\renewcommand{\paragraph}[1]{\noindent\textbf{#1}}

\usepackage{amsmath}
\usepackage{amssymb}
\usepackage{mathtools}
\usepackage{amsthm}
\usepackage{multirow}

\usepackage[capitalize,noabbrev]{cleveref}

\theoremstyle{plain}

\theoremstyle{definition}

\theoremstyle{remark}

\usepackage[textsize=tiny]{todonotes}

\usepackage[binary-units]{siunitx}

\icmltitlerunning{Cache Me If You Must: Adaptive Key-Value Quantization for Large Language Models}

\begin{document}

\twocolumn[
\icmltitle{Cache Me If You Must: Adaptive Key-Value Quantization \\ for Large Language Models}



\icmlsetsymbol{equal}{*}

\begin{icmlauthorlist}
\icmlauthor{Alina Shutova}{equal,hse}
\icmlauthor{Vladimir Malinovskii}{equal,yandex,hse}
\icmlauthor{Vage Egiazarian}{equal,ista}
\icmlauthor{Denis Kuznedelev}{yandex}
\icmlauthor{Denis Mazur}{sberdevices,mipt}
\icmlauthor{Nikita Surkov}{tbank}
\icmlauthor{Ivan Ermakov}{mipt}
\icmlauthor{Dan Alistarh}{ista}
\end{icmlauthorlist}

\icmlaffiliation{ista}{ISTA}
\icmlaffiliation{hse}{HSE University}
\icmlaffiliation{mipt}{MIPT}
\icmlaffiliation{tbank}{T-Bank}
\icmlaffiliation{sberdevices}{SberDevices}
\icmlaffiliation{yandex}{Yandex}

% \icmlcorrespondingauthor{Alina Shutova}{first1.last1@xxx.edu}
\icmlcorrespondingauthor{Dan Alistarh}{dan.alistarh@ist.ac.at}

\icmlkeywords{Machine Learning, ICML}

\vskip 0.3in
]



\printAffiliationsAndNotice{\icmlEqualContribution} % otherwise use the standard text.

\begin{abstract}\vspace{-2px}
 Efficient real-world deployments of large language models (LLMs) rely on Key-Value (KV) caching for processing and generating long outputs, reducing the need for repetitive computation. For large contexts, Key-Value caches can take up tens of gigabytes of device memory, as they store vector representations for each token and layer. Recent work has shown that the cached vectors can be compressed through quantization, pruning or merging, but these techniques often compromise quality towards higher compression rates. In this work, we aim to improve Key \& Value compression by exploiting two observations: 1) the inherent dependencies between keys and values across different layers, and 2) high-compression mechanisms for internal network states. We propose AQUA-KV, an adaptive quantization for Key-Value caches that relies on compact adapters to exploit existing dependencies between Keys and Values, and aims to ``optimally'' compress the information that cannot be predicted. AQUA-KV significantly improves compression rates, while maintaining high accuracy on state-of-the-art LLM families. On Llama 3.2 LLMs, we achieve near-lossless inference at 2-2.5 bits per value with under $1\%$ relative error in perplexity and LongBench scores. AQUA-KV is one-shot, simple, and efficient: it can be calibrated on a single GPU within 1-6 hours, even for 70B models.

\end{abstract}

\vspace{-2.5em}
\section{Introduction}\label{sect:intro}
\vspace{-3px}

\begin{figure}[ht]
    \centering
    \includegraphics[width=0.98\linewidth]{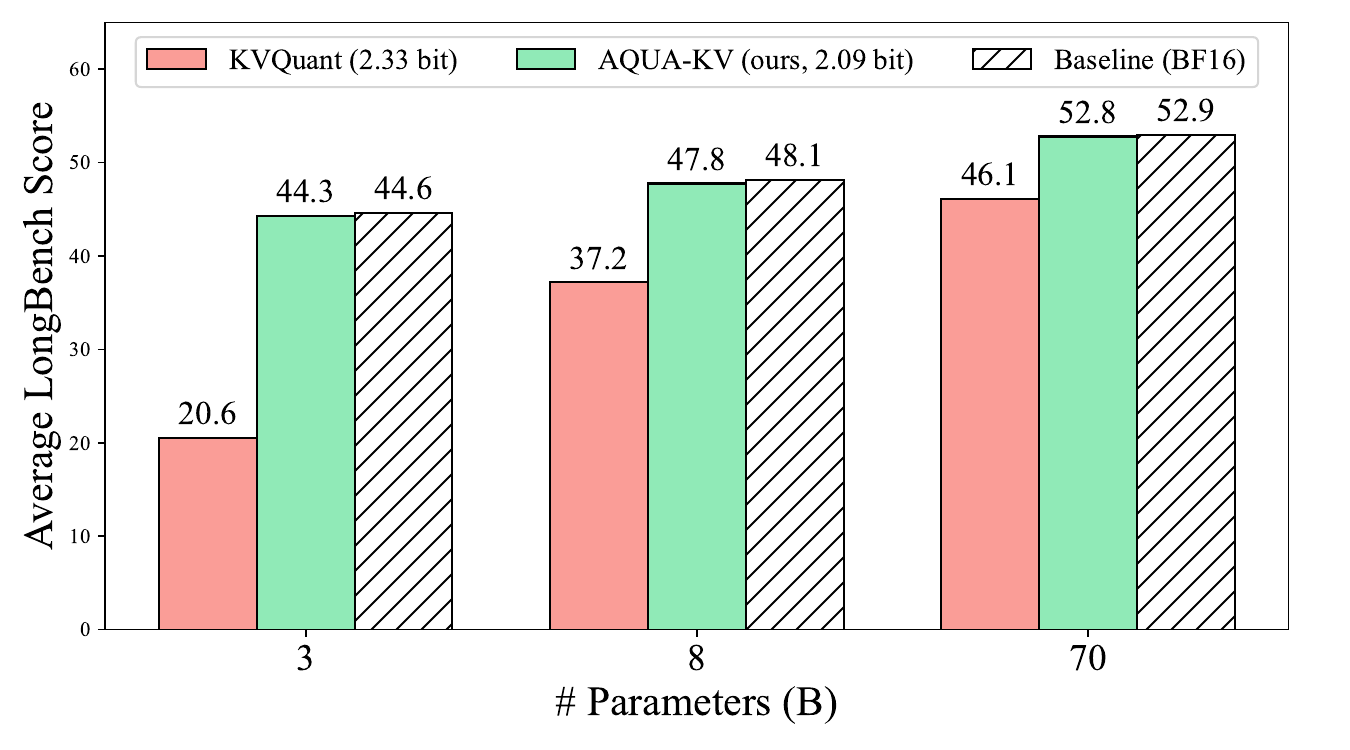}
    \vspace{-14px}
    \caption{
    Comparison of AQUA-KV to alternative Key-Value Cache compression methods for Llama 3.x models in terms of average LongBench score on 14 english tasks (see Section~\ref{sect:experiments}).}
    \label{fig:teaser}
    \vspace{-18px}
\end{figure}

% \begin{enumerate}
%     \item LLMs exist and are important
%     \item LLM inference uses KV caches
%     \item applications demand long sequences, long sequences mean very large KV caches (give specific numbers, e.g. llama3.1 8B model 16GB(?) 128k cache 42GB per user)
%     \item methods to compress KV caches exist: quantization, token pruning, channel pruning, etc
%     \item to compress is to understand; there is a structure in KV caches that we're not using. We can exploit mutual information between KV caches, only store what cannot be predicted
%     \item our approach - combine mutual information (via small predictors) with modern VQ + RHT (HIGGS / data-free QUIP). Despite having calibration phase, our algorithm uses closed form solutions and takes TODO hours for THIS LARGE model on a single  gpu.
% \end{enumerate}

Large Language Models (LLMs) are revolutionizing natural language processing, but come with major computational costs,
in particular due to the input-quadratic complexity of attention-based Transformer models~\cite{vaswani2017attention}.
To achieve faster inference and avoid wasteful recomputation of  attention scores during autoregressive generation, \textit{KV caching} is typically employed, where keys and values are saved for later use.
% LLM inference uses KV caches; KV caches are as follows.
Unfortunately, KV-caching comes with its own pitfalls: 
 KV caches are large, especially when handling long sequences~\cite{bai2023longbench, xiao2023efficient}.
Thus, the memory footprint of a full-length Key-Value cache can reach tens of gigabytes of device memory, sometimes more than the model itself\footnote{For the popular Llama 3.2 3B model~\cite{dubey2024llama} with a maximum context length of $2^{17}$ tokens (${\approx}$131K), the 16-bit cache takes up $\SI{15}{\giga\byte}$ per sequence. For Llama 3.1 70B and Qwen 2.5 72B, it is $\SI{42.9}{\giga\byte}$ per sequence.}.
% For instance, considering e.g. Llama-3.1~70B with context of 128k tokens, the cache would take $\SI{42}{\giga\byte}$ per user.
This massive memory consumption increases the cost of deployment, and also slows down inference, as the whole process can become memory-bound for large caches~\cite{hooper2024kvquant}.

Previous work has proposed methods to compress KV caches using various methods such as quantization and pruning~\cite{li2024survey}, that can significantly reduce the memory footprint of KV caches.
Yet, as we increase the degree of compression, e.g. $2$ bits per value, existing compression techniques begin to lose significant information, resulting in poor accuracy~\cite{li2024scbench}.
% In order to make these compression rates feasible, it is crucial to gain deeper understanding of the underlying data.

In this work, we aim to improve KV cache compression by taking advantage of the inherent structure and dependencies in the cache tensors.
Specifically, we analyze the Key-Value cache behavior for state-of-the-art LLMs and find several strong inter-dependencies 1) between cached vectors from adjacent layers, but also 2) between the keys and values within one layer.

Starting from these observations, we formulate a practical compression algorithm that explicitly leverages these inter-dependencies, by training compact linear predictors that capture mutual information between cache components.
Further, to offset prediction errors, we use data-free vector quantization to achieve a superior compression-accuracy trade-off for the same bit-width.
Our method requires only minimal calibration, is compatible with arbitrary quantization schemes and can be further combined with  orthogonal compression techniques such as pruning.

In summary, our contributions are as follows:
\begin{enumerate}[leftmargin=*]
    \vspace{-10px}
    \item We analyze the structure of key-value caches in modern LLMs and highlight several sources of mutual information that can be leveraged for compression.
    \vspace{-5px}\item We propose \methodname{} --- a novel compression framework that 
    exploits inter- and intra-layer dependencies to improve quantization accuracy.
    \methodname{} works in one shot, based on a lightweight calibration procedure, and shows competitive size-accuracy trade-offs. Additionally, it is compatible  with arbitrary quantization techniques, and can be combined with additional compression, such as pruning.
    \vspace{-5px}\item We validate the effectiveness of \methodname{} on modern LLM families in terms of both perplexity and zero-shot accuracy on long-range benchmarks, where AQUA-KV significantly improves accuracy across model types and bitwdiths, particularly for 2-bit compression.
    \vspace{-5px}\item We test AQUA-KV compatibility with various quantization and pruning schemes, from simple uniform quantizatin, to modern data-free vector quantizers~\cite{malinovskii2024pushing}\,and\,hybrid\,quantization\,\&\,pruning regimes.\hspace{-2px}
    \vspace{-15px}\item We develop a reference implementation for \methodname{} calibration and inference, which is available online\footnote{\url{https://github.com/goodevening13/aquakv}}.
\end{enumerate}\vspace{-10px}

\vspace{-8px}
\section{Background and Related Work}\label{sect:background}
% One of them is efficient attention inference, which is quadratic in its time complexity.
% Modern decoder-only LLMs typically use autoregressive inference, i.e. they generate tokens in several passes, adding generated tokens to the prompt.
% In order to avoid costly recomputation of attention between existing tokens on every iteration \textit{KV caching} is employed.
% Storing keys and values in memory brings significant memory overhead.

\subsection{KV-Cache Compression}\label{sect:background_kvcache_quantization}

So far, the main focus of work on LLM compression has been on the \emph{weight quantization}, e.g.~\cite{frantar2022gptq, lin2023awq, tseng2024quip, egiazarian2024extreme}. 
Recently, there has been a growing demand for KV-cache compression, especially in tasks requiring long contexts.
The dynamic nature of KV-caching poses unique challenges: 
while for weights it is acceptable to use ``slow but accurate'' compression such as codebook-based methods~\citep{tseng2024quip, egiazarian2024extreme}---given that weights are only decoded at inference time---for the KV-cache, \emph{both compression and decompression speeds matter}, since we are dynamically adding new entries to the cache as well as decoding them at inference time.
Another issue is posed by the inherent structure in caches, in particular, the existence of attention sinks~\cite{xiao2023efficient} and large outlier values~\cite{liu2024kivi, hooper2024kvquant, liu2024minicache}, which may not be present in the weights.
Next, we detail the main approaches for KV-cache quantization.
\vspace{-2px}

KV-Cache quantization approaches can be roughly categorized based on quantization granularity and error handling.

% With increased demand, tailored algorithms specifically for KV caches have been developed.
% Mainly, the methods fall into the following categories:
% \begin{enumerate}
%     \item KV Cache Quantization
%     \item KV Cache Cross-Layer Merging
%     \item KV Cache Pruning
% \end{enumerate}
% \begin{enumerate}
%     \item mention KIVI, KVQuant, any other baselines we have
%     \item mention pruning; position us as orthogonal to pruning
%     \item mention layer merging; position us as a more fine-grained approach
% \end{enumerate}

% \todo[inline]{New text starts here.}

\vspace{-2px}
\paragraph{Quantization Granularity.}
Several approaches have been developed with varying quantization granularities. For instance, ZipCache~\cite{he2024zipcache} and WKVQuant~\cite{yue2024wkvquant} implement channel-separable token-wise quantization, while KVQuant~\cite{hooper2024kvquant} and KIVI~\cite{liu2024kivi} employ a hybrid approach, using per-channel quantization for key tensors while applying per-token quantization for value tensors. QJL~\cite{zandieh2024qjl} introduces a specialized JL transform for key tensors combined with per-token quantization for value tensors. Methods like MiKV~\cite{yang2024mikv}, QAQ~\cite{dong2024qaq}, and SKVQ~\cite{duanmu2024skvq} employ variable bit widths to  balance  accuracy and memory reduction.

\vspace{-2px}
\paragraph{Error Handling.}
Among the strategies used to address quantization errors, GEAR~\cite{kang2024gear} compensates for errors using a low-rank matrix; to handle  outliers, which can significantly impact model performance, IntactKV~\cite{liu2024intactkv} maintains full precision for outlier values. QuaRot~\cite{ashkboos2024quarot} transforms weight matrices using Hadamard orthogonal matrices to ``smoothen'' quantization outliers without affecting model output. Palu~\cite{chang2024palu} compresses KV cache through low-rank projection, while ZDC~\cite{zhang2024zdc} aims to eliminate compression overhead through a novel zero-delay compression scheme.
Most methods maintain a window of recent historical KV cache in full precision to preserve accuracy. 
%This approach has proven effective in maintaining model performance while achieving significant memory reduction.

\vspace{-2px}
\paragraph{Cross-Layer Merging.}
An alternative compression approach, which has been relatively less investigated, has been to define layer groups, and keep a single KV-Cache per layer group, reusing the one cache across all layers in the group.
KVSharer~\cite{yang2024kvsharer} observes that, surprisingly, sharing the caches that are \emph{most dissimilar}, by the Euclidean distance, performs better than sharing similar ones. (This is identified using a calibration dataset.)
% While this approach facilitates efficient inference due to lack of additional computations, directly reusing KV Caches is suboptimal because of differences between layers.
% The development of the above approach is to maintain a shared representation of caches per group of layers and use an additional transform to get layer cache during inference.
Further, MiniCache~\cite{liu2024minicache} merges pairs of layers, and stores a common interpolated directional vector and token-wise scalar scales.
Additionally, outlier tokens are retained to increase accuracy.
During inference, keys and values of the two layers are restored from the shared vector representation using saved scales.

\vspace{-2px}
\paragraph{KV Cache Pruning.}
Pruning methods aim to determine unnecessary parts of KV Caches and either evict them entirely, or offload to cheaper memory (e.g. CPU).
Current research focuses on token-level pruning, e.g. determining which tokens should be discarded or offloaded.
They can be split into static methods and dynamic methods. 

\emph{Static methods} use predefined position heuristics to determine important tokens.
An example is Fastgen~\cite{ge2023model}, which employs knowledge of attention structure in attention heads, acquired during prefill, to identify one of the tailored attention structures for each head.
This knowledge is then used during inference to efficiently evict unnecessary tokens.
StreamingLLM~\cite{xiao2023efficient} notices that initial tokens (attention sinks) and recent tokens consistently exhibit high importance.
The method retains several tokens in the beginning and a number of recent tokens, thus maintaining constant-size KV cache, facilitating deployment in memory-constrained scenario.

\emph{Dynamic methods} calculate importances dynamically, typically using runtime information about attention distribution. H2O~\cite{zhang2023h2o} calculates token importances dynamically during inference using accumulated attention scores.
It selects the least important tokens during each forward pass (if the cache is larger than desired) and thus maintains constant cache size.
SnapKV~\cite{li2024snapkv} calculates token importances using attention features;
unlike H2O, SnapKV thresholds token importances, allowing prompt compression during prefill.

For our work we choose to conduct experiments using H2O pruning as it is a well-established plug-and-play method. However, we emphasize that our method is orthogonal to pruning and AQUA-KV can be combined with any cache eviction strategy.
% In this context, AQUA-KV, lies at the intersection of Quantization and Cross-Layer Merging approaches.
% On one hand, the method can be viewed as a more fine-grained approach to Layer Merging by training a small supplementary model, referred to as \textit{predictor}, to grasp inter-layer dependencies of KV cache.
% Moreover, AQUA-KV leverages an advanced quantization technique~\citep{higgs} to efficiently store quantized prediction residuals, so it can also be seen as a new type of quantization technique.
AQUA-KV  integrates both Quantization and Cross-Layer Merging strategies.
It can be regarded as more advanced Layer Merging approach via 
training a small supplementary model, referred to as \textit{predictor}, capturing inter-layer dependencies of KV caches. 

% \textcolor{gray}{\lipsum[5-6]}

\begin{figure*}[t]
    \begin{minipage}{.48\textwidth}
        % \vspace{-10px}
        \includegraphics[width=0.95\linewidth,height=149px]{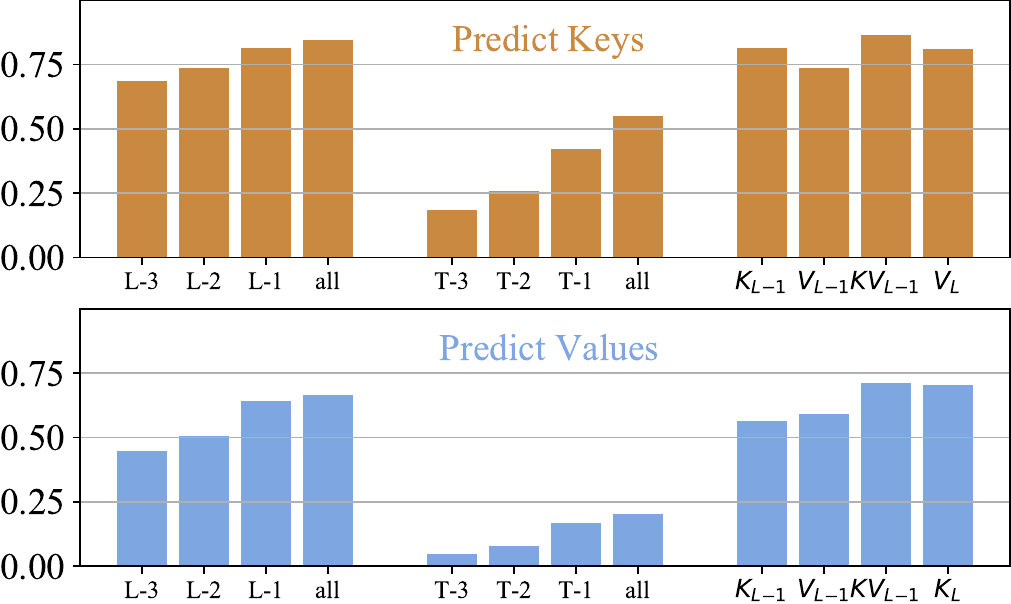}
        \vspace{-9px}
        \caption{
        Mean Explained Variance Ratios by linear probes from previous blocks (L), tokens (T) and role on Llama-3.2-3B.
        }\label{fig:3.1_predictability}
        \vspace{-14px}    
    \end{minipage}%
    \hfill
    \begin{minipage}{.48\textwidth}
        % \vspace{-10px}
        \includegraphics[width=\linewidth]{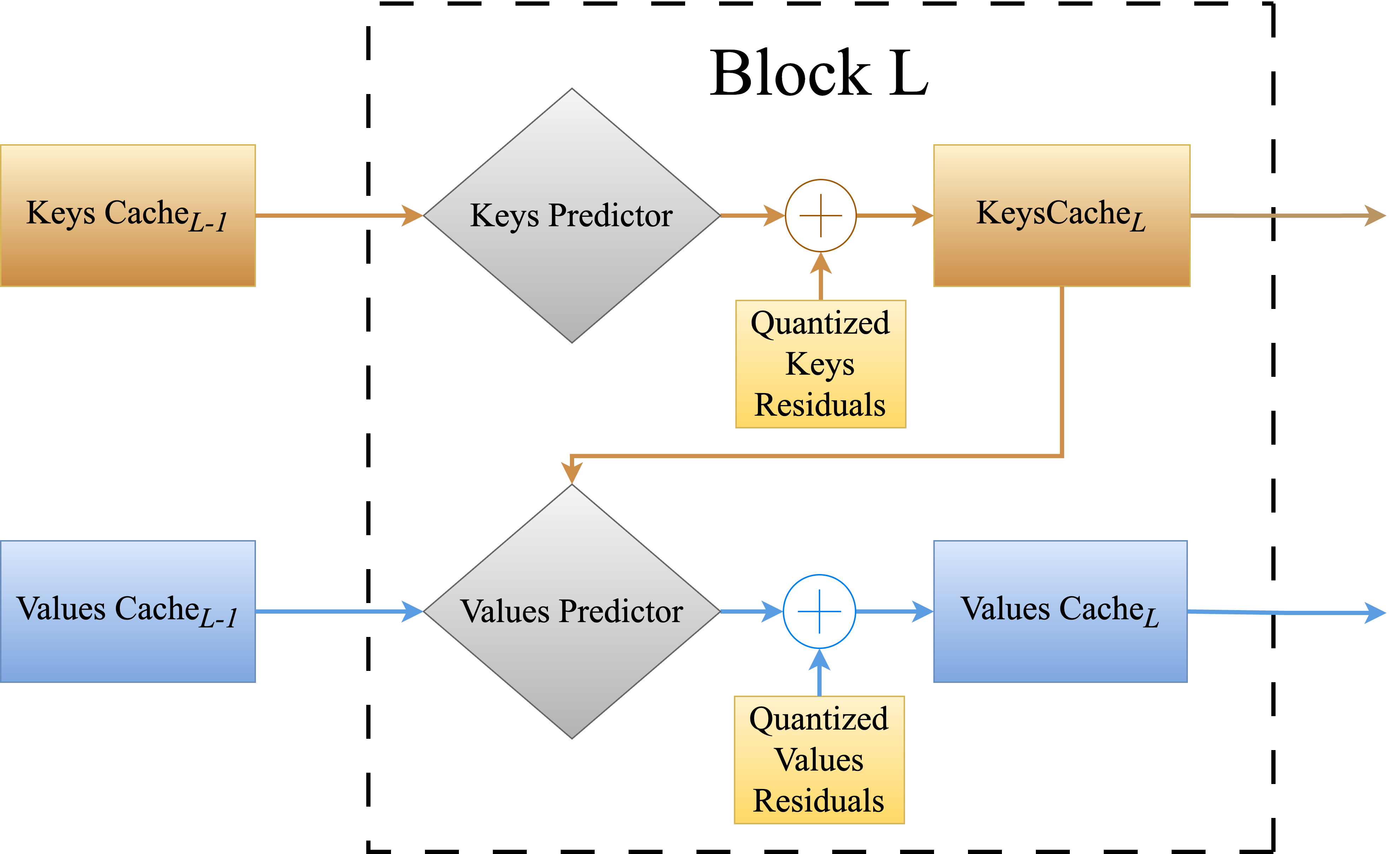}
        \vspace{-17px}
        \caption{An intuitive scheme of the AQUA-KV inference.\\Only the quantized residuals are saved for each block.
        }
        \label{fig:3.1_scheme}
        \vspace{-16px}    
    \end{minipage}%
\end{figure*}

\paragraph{Vector Quantization}\label{sect:background_rq} 
 recently emerged as a popular option for LLM quantization~\citep{egiazarian2024extreme, tseng2024quipsharp, van2024gptvq} since it allows to jointly quantize multiple individual model dimensions and can lead to state-of-the-art accuracy-vs-compression~\citep{tseng2024qtip, malinovskii2024pv}. 
However, the extremely large computational cost of encoding makes such methods impractical in the context of KV-cache compression. 

We resolve this issue, and leverage the power of VQ in an efficient way, by adapting the HIGGS weight quantization technique~\citep{malinovskii2024pushing} to KV-Cache compression. 
HIGGS combines group-wise vector quantization with a Randomized Hadamard Transformation (RHT): the RHT projects  the original values onto a ``rotated'' space, where they will be normally-distributed. In the quantization step, the rotated values are grouped and rounded to the nearest points on a lattice, which is specifically optimized for accurate quantization of normally-distributed vectors. 
This allows HIGGS to achieve \emph{fast data-free weight quantization}. The technique requires several modifications to be applied to Key-Value caches, which we detail in Section~\ref{sect:experiments_main}. 
Finally, our approach is conceptually related to Residual Vector Quantization (RVQ)~\citep{gray1992vector}, but using learned predictors instead of standard quantization. 

\paragraph{Linearity between adjacent layers.}\label{sect:linearity} Prior work \cite{razzhigaev2024your} has shown that there is a almost linear relationship between activations in sequential layers in transformer language models due to the presence of skip connections. 
We leverage this property for the design of KV-cache predictors.

% \begin{enumerate}
%     \item mention basic VQ , explain why it works, mention that it requires expensive KMeans
%     \item mention RHT plus lattices to avoid KMeans; mention JL as similar to RHT
%     \item mention Residual Vector Quantization; position us as philosophically similar, but with predictors
% \end{enumerate}

% \textcolor{gray}{\lipsum[7-8]}

\vspace{-0.7em}
\section{Method}\label{sect:method}

The core idea of AQUA-KV is to leverage inter-dependencies between consecutive KV-caches to improve compression. \textbf{For this, we train compact predictors that ``guess'' the value of a Key \& Value pair using other cache entries, then quantize the residual information that could not be predicted.} This way, we only store the information that cannot be recovered from other sources.

In Section~\ref{sect:method_dependencies}, we analyze the dependencies between various KV cache components to determine the type of predictors that achieves the best size-accuracy trade-off. In Section~\ref{sect:method_algorithm}, we formulate a practical one-shot algorithm that fits these predictors for use in KV cache compression. Finally, in Section~\ref{sect:method_implementation} we describe a number of important implementation details for using AQUA-KV in practice.

\subsection{Analysis of Inter-Layer Dependencies}\label{sect:method_dependencies}

The efficacy of our approach depends on choosing which types of inter-dependencies to exploit. To make this choice, we analyze the dependencies between cached vectors at different layers or tokens, and between (key \& value) vectors within the same layer. Note that we do not expect the values in these vectors to be equal or even numerically close---indeed, a simple examination shows that this is not the case. Instead, we look for consistent dependencies between these components that can be extracted with simple models.

To measure dependencies between vectors at nearby layers, we adopt an approach similar to probing~\cite{alain2016understanding}: we train \textit{linear} ``probe'' models whose goal is to predict the contents of a particular cache component, e.g. $i$-th layer keys or values, based on inputs from various sources. As potential sources, we consider the previous layer keys and values, adjacent tokens, and different vector types (i.e. using the keys to predict values and vice versa).

We can then measure the relative prediction error for such probes, based on small amounts of calibration data. More specifically, we measure the explained variance ratio to account for unequal scales of keys and values between layers. Intuitively, if a predictor captures 90\% of the variance, it means that the subsequent quantization only needs to capture the remaining 10\% of variance. For compression methods that are scale-independent~\footnote{Formally, a scale-independent quantizer $Q(\cdot)$ satisfies $\forall X, \forall \alpha {>} 0, \; ||Q^{-1}(Q(\alpha X)) - \alpha X||_2 = \alpha ||Q^{-1}(Q(X)) - X||_2$. Modern quantizers satisfy this due to the use of scales.}\cite{horvath2023stochastic} this would mean that the resulting quantization will also have roughly 10 times smaller error.

Note that not all predictors will be practical for Key-Value compression. For instance, if a predictor uses a subsequent block or future token KV vectors as inputs for the current ones, it would be difficult to inference the model with such KV cache as it goes against the order in which the blocks execute. Hence, we consider the following:
\vspace{-5px}\begin{itemize}[leftmargin=*]
    \vspace{-5px}\item \textbf{Previous blocks:}\,same token\,vectors\,for\,-1,\,-2,\,-3 blocks;
    \vspace{-5px}\item \textbf{Previous tokens:} same layer, -1, -2, -3 previous tokens;
    \vspace{-5px}\item \textbf{Different \hspace{-1px}role:}\,using\,keys\,to predict\,values\,and\,vice\,versa.
\end{itemize}\vspace{-5px}

We report these errors in Figure~\ref{fig:3.1_predictability}. For comparison, 1-bit and 2-bit quantizers usually explain 0.75 and 0.89 variance, respectively. Intuitively, if a probe can predict the $L$-th block keys with the same relative error as the 1- or 2-bit quantizer, it means that using the probe can ``save'' approximately this many bits per value when quantizing residuals. While this is not a strict guarantee, we found that it holds well for real-world LLMs, as can be seen in Sections~\ref{sect:experiments_first} and~\ref{sect:experiments_main}.

We train linear probes for Llama-3.2-3B Key-Value cache on a sample of RedPajama~\cite{together2023redpajama} sequences, then evaluate relative error on hold-out sequences from the same source. For readability, the detailed experiment configuration is deferred for Appendix~\ref{app:method_details}.

The findings in Figure~\ref{fig:3.1_predictability} demonstrate strong dependencies between several cache components. For attention keys, using just one previous layer already achieves errors similar to 2-bit quantization. For values, the dependency on previous layer is also strong: though less accurate than for keys, the previous layer values consistently explains more than half of the variance for the same token. We attribute this strong dependency to the fact that transformer architecture is residual, and therefore, adjacent hidden states are only off by a single transformer layer. Since Key-Value representations are constructed as linear projections of adjacent residual hidden states, they are also interdependent.

More distant layers are also predictive of the current layer, but the dependency quickly deteriorates with the distance. More importantly, there is almost no difference between using multiple past layers and just the previous layer, which allows us to simplify the algorithm. See Appendices~\ref{app:extra_figure2_results}~\&~\ref{app:additional_results_4.1} for additional exploration and downstream results.

We also observe strong dependencies between keys and values within the same layer. This is also not unexpected, as these vectors are linear down-projections from the same input vector within the attention layer. The reason why there is no additional information between matching keys/values is that in modern LLMs the key/value dimensions are both significantly smaller than the input vector dimension due to Grouped Query Attention~(GQA, \citealt{ainslie2023gqa}).

In contrast, we found relatively little predictive power from past \textit{tokens}. While \textit{there is} mutual information between adjacent tokens, using a single previous token is not enough to capture this dependency, and using many multiple tokens or a more expressive probe would make the resulting KV cache compression inefficient.

Some of our observations above correlate with the results from \citet{liu2024minicache}, who were able to completely ``share'' some of the cache entries. However, such direct merging is a drastic approach, that can lead to significant accuracy loss. From the perspective of Figure~\ref{fig:3.1_predictability}, this is because the information recovered by a probe is not enough by itself, never exceeding the equivalent of a 2-bit quantizer. In contrast, we employ a more fine-grained approach that combines predictors with residual quantization that can achieve more accurate compression.

% According to our analysis, this is attributed to the fact that the mutual information varies between layers and is often not enough by itself, necessitating a more careful approach.

% \begin{enumerate}
%     \item Observation - KV layers are highly interdependent, but not similar. There exists a simple transition between adjacent layers that captures most of the information
%     \item Maybe some demo plot to showcase it
%     \item The information is substantial, but often not enough to merge layers completely without significant loss in accuracy
%     \item Core idea of our approach - use compact (linear) predictors to capture mutual information, compress only what cannot be predicted
% \end{enumerate}

\begin{figure}[h]
\vspace{-10px}
\begin{algorithm}[H]
\caption{AQUA-KV Calibration}
\label{alg:method_calibration}
\begin{algorithmic}[1]
\REQUIRE model, data, quantization method $Q$

\STATE $\mathbf{X} \gets \texttt{model.input\_embeddings(data)}$
\STATE $\mathbf{K}_\text{old}, \mathbf{V}_\text{old} \gets \emptyset$
\STATE $\texttt{predictors} \gets \{\}$
\FOR{$i = 1, \dots, \texttt{model.num\_layers}$}
    \STATE $\texttt{block} \gets \texttt{model.transformer\_layers}[i]$
    \STATE $(\mathbf{K}, \mathbf{V}) \gets \texttt{block.get\_attention\_kv}(\mathbf{X})$
    \STATE $\mathbf{X} \gets \texttt{block}(\mathbf{X})$
    
    \vspace{3 px}
    % \STATE \# First iteration
    \IF{$\mathbf{K}_\text{old} = \emptyset$ \AND $\mathbf{V}_\text{old} = \emptyset$}
        \STATE $\mathbf{K}_\text{old}, \mathbf{V}_\text{old} \gets \mathbf{K}, \mathbf{V}$
        \STATE \textbf{continue}
    \ENDIF
    \vspace{3 px}
    
    \STATE \# Predict keys from past keys
    \STATE $f_\text{key} \gets \arg\min_f \| f(\mathbf{K}_\text{old}) - \mathbf{K} \|_2^2$
    \STATE $\mathbf{K}_{\text{residue}} \gets Q^{-1}(Q(\mathbf{K} - f_\text{key}(\mathbf{K}_\text{old})))$
    \STATE $\mathbf{K}_{\text{rec}} \gets f_\text{key}(\mathbf{K}_\text{old}) + \mathbf{K}_{\text{residue}}$

    \vspace{3 px}
    \STATE \# Predict values from past values and current keys
    \STATE $f_\text{value} \gets \arg\min_f \| f([\mathbf{V}_\text{old}; \mathbf{K}_{\text{rec}}]) - \mathbf{V} \|_2^2$
    
    \STATE $\mathbf{V}_{\text{residue}} \gets Q^{-1}(Q(\mathbf{V} - f_\text{value}([\mathbf{V}_\text{old}; \mathbf{K}_{\text{rec}}])))$
    \STATE $\mathbf{V}_{\text{rec}} \gets f_\text{value}([\mathbf{V}_\text{old}; \mathbf{K}_{\text{rec}}]) + \mathbf{V}_{\text{residue}}$
    
    \vspace{3 px}
    \STATE $\texttt{predictors[i]} = (f_\text{key}, \, f_\text{value})$
    \STATE $\mathbf{K}_\text{old}, \mathbf{V}_\text{old} \gets \mathbf{K}_{\text{rec}}, \mathbf{V}_{\text{rec}}$
\ENDFOR
% \vspace{3 px}
\STATE \textbf{return} \texttt{predictors}
\end{algorithmic}
\end{algorithm}
\vspace{-30px}
\end{figure}

\vspace{-3px}
\subsection{The AQUA-KV Algorithm}\label{sect:method_algorithm}
\vspace{-3px}

Based on the findings from the previous section we design a cache compression algorithm that leverages the structure of Key-Value cache to improve compression.

We choose the following predictor configuration: (1) we use the previous layer keys to predict the subsequent keys; and (2) we use both previous layer values and current layer keys to predict values. Note that we do \textit{not} use the dependency between same layer K and V vectors in the $V_L{\rightarrow}K_L$ direction: this is because we cannot predict in both directions simultaneously during inference, and we found that the values are overall harder to predict (see Figure~\ref{fig:3.1_predictability}), so we opted to improve value predictor. We also purposefully leave out other possible input sources, such as more distant past layers. While using these layers can slightly improve reduce error, it would also make the predictors themselves larger and more compute-intensive.

The way AQUA-KV trains those predictors is also different from Section~\ref{sect:method_dependencies}. This is because, in practical KV cache compression, the predictors do not have access to ground-truth past KVs during inference. Instead, they can only use reconstructed (de-quantized) past key and value vectors.
To account for this discrepancy, AQUA-KV trains predictors sequentially, one transformer layer at a time. Each subsequent set of predictors is trained using reconstructed cache entries as inputs, reflecting the way these predictors are used during inference. The first layer key-value cache is compressed as is, and each subsequent layer trains using previous layer key-value \textit{reconstructions} as inputs. 
The full calibration procedure is described in Algorithm~\ref{alg:method_calibration}. 
%In our main setup, the key predictor uses previous layer keys, and the value predictor uses previous layer values and reconstructed \textit{current layer keys}, aligning with our earlier findings in see Section~\ref{sect:method_dependencies}. We provide an ablation over alternative input sources in Section~\ref{sect:experiments_ablation}.

Here, $Q(\cdot)$ and $Q^{-1}(\cdot)$ denote quantization and de-quantization operators. Our algorithm is agnostic to the choice of $Q(\cdot)$: simple uniform min-max quantization or any advanced method can be used~\cite{frantar2022gptq, malinovskii2024pushing}. The notation ${\arg\min}_f || f(X) - Y||_2^2$ fits a linear regressor to a given problem.
By default, we use simple linear regression for all predictors, which allows for a closed-form training and is easy to use during inference. However, our algorithm can accommodate any other regressor type. In~\ref{sect:experiments_first} we consider alternative  algorithms: reduced-rank regression~\cite{reinsel1998multivariate} and MLPs.

\vspace{-2px}
\textbf{Computational and memory overhead.} In addition to the cache entries, AQUA-KV also needs to store the trained predictors during inference. This requires additional computation, and slightly increases the memory footprint.
However, this overhead is small in practice: this is largely because modern LLMs use GQA~\cite{ainslie2023gqa}---a popular attention variant that uses fewer key-value heads than queries. For AQUA-KV, this means that the predictors contain significantly less parameters and require orders of magnitude less computation. For instance, For Llama 3.x 70B and Qwen 2.5 72B, inferencing AQUA-KV predictors for a token requires at least $500\times$ less floating point operations than running the base model for the same token (see Section~\ref{sect:experiments_main}).

% . As such, the practical overhead of those predictors is small, as we show in Section~\ref{sect:experiments}. However, to accommodate edge cases where the predictor footprint needs to be smaller, we also consider compressed (low-rank or quantized) predictors in Section \textbf{TODO}.

\vspace{-2px}
\textbf{Efficiency and limitations.} Algorithm~\ref{alg:method_calibration} is designed to work as a single-pass calibration procedure. It trains predictors separately, one at a time. This allows our algorithm to run efficiently even on low-end hardware. For instance, calibrating the full set of AQUA-KV predictors for a Llama-3.1-70B model takes up 4 hours on a single GPU and takes up at most 16GB VRAM.
AQUA-KV is designed as a simple and lightweight algorithm that can be easily extended. As such, we deliberately forego more complex techniques such as global predictor fine-tuning or dynamic bitwidth. We discuss these possible extensions in Section~\ref{discussion}.

\vspace{-2px}
\subsection{Implementation Details}\label{sect:method_implementation}\vspace{-2px}

Finally, we describe several practical  details necessary for efficient implementation of our approach, whose experimental validation is given in Section~\ref{sect:experiments_first}.

\vspace{-1px}
\textbf{Backbone quantization.} To validate the generality, we couple AQUA-KV with two different quantization schemes: Quanto~\citep{optimum-quanto} (simple round-to-nearest quantization with absmax normalization) and the more advanced HIGGS~\citep{malinovskii2024pushing}.

\vspace{-1px}
\textbf{Relation between AQUA-KV and Attention Sinks.} It is well-known that modern LLMs tend to form attention ``sinks'' --- tokens that have extremely high attention scores while being semantically unimportant~\cite{xiao2023efficient}. Several prior works in KV compression propose special treatment for such attention sinks, such as keeping them in higher precision~\cite{hooper2024kvquant} or introducing synthetic sinks~\cite{xiao2023efficient,chen2025prefixquanteliminatingoutliersprefixed}.

When evaluating AQUA-KV, we found that attention sinks are indeed important. Compressing attention sinks poorly can affect the behavior of attention heads on other tokens and, hence, change the input/output distribution for the learned predictors.
The AQUA-KV calibration algorithm computes input keys and values without accounting for quantization error (Alg.~\ref{alg:method_calibration} L5-7), as doing otherwise would significantly increase calibration time. Hence, we found that AQUA-KV benefits from keeping the first few tokens uncompressed, similarly to how they are treated in KVQuant~\cite{hooper2024kvquant}. For fair comparison, we keep the first 4 tokens uncompressed for both AQUA-KV and baselines without predictors, and explore this in more detail in Table~\ref{tab:app_extra_analysis_for_4.1} in suplementary materials.

% \vspace{-1px}
\textbf{Positional embeddings.} Most LLMs apply Rotary Positional Embeddings (RoPE,~\citealt{su2021roformer}) to attention keys (but not values). This raises a natural dilemma about whether to apply predictors and quantizers before or after quantization. In our analysis, we found that linear predictors are beneficial in either case, but they offer better accuracy in pre-RoPE compression. We attribute this to the fact that the optimal predictor for post-RoPE compression needs to be rotation-equivariant, and the simple linear models we use are not. As for the backbone quantization, we found that the uniform quantizer (Quanto) works slightly better post-RoPE, while HIGGS works equally well in both cases due to its use of the Hadamard transform. We provide experimental validation for these claims in Appendix~\ref{app:additional_results_4.1}.

% \vspace{-1px}
\textbf{Per-token and per-channel compression.} We use per-token quantization for both schemes. While some prior works suggest that keys are better quantized per-channel~\cite{liu2024kivi,hooper2024kvquant} due to different outlier structure, we found that this is not necessary when quantizing predictor residuals. We use per-token compression as it is easier to implement, and we ablate this choice in Appendix~\ref{app:additional_results_4.1}. For baselines, we always follow the quantization axes suggested in their respective papers.

% \vspace{-1px}
\textbf{Inference Algorithm.} Finally, we explain how these design choices combine to the full AQUA-KV inference procedure. Both AQUA-KV and all our baselines maintain a small recent token buffer (up to $r{=}128$ tokens for all setups) that are originally stored without compression. This buffer has two positive effects: it improves accuracy on recent tokens and allows for efficient parallel processing.
When the buffer is filled, its contents are quantized incrementally from the first layer to the last, in the same order as during calibration.

We define this procedure formally in Alg.~\ref{alg:method_inference}.
Due to the sequential nature of our approach, we only need to materialize (de-quantize) a single layer at a time, which can be further improved with chunking. Note that the proposed scheme is seamlessly compatible with model parallelism, offloading, speculative decoding, merging multiple sequences and other popular LLM inference techniques. Furthermore, as we show in Section~\ref{sect:experiments_third}, our approach can be used in tandem with pruning techniques, such as H$_2$O~\cite{zhang2023h2o}.

\vspace{-0.7em}
\section{Experiments}\label{sect:experiments}

\vspace{-1px}
To test the real-world effectiveness of AQUA-KV, we apply it to modern LLMs in three setups: 1) in Section~\ref{sect:experiments_first}, we analyze the impact of individual components of our method and verify the design choices 2) Section~\ref{sect:experiments_main} evaluates on a broader range of models and compression rates; 3) finally, Section~\ref{sect:experiments_third} explores how our approach combines with other popular KV cache compression strategies.

\vspace{-1px}
Across all three sections, we use the same calibration and evaluation protocol. We use a sample from RedPajama~\cite{together2023redpajama} dataset for calibration: namely, 256 sequences of 8192 tokens sampled at random. We use 32 of those sequences as holdout for hyperparameter selection and the remaining 224 are used to train the predictors themselves. We use two popular evaluation metrics: WikiText-2~\cite{wikitext103} perplexity and LongBench~\cite{bai2023longbench}.

\vspace{-1px}
When evaluating perplexity, we adopt the same approach as in prior quantization works~\cite{frantar2022gptq,lin2023awq,egiazarian2024extreme}, with one exception: instead of processing sequences (of length 8192) in parallel, we encode them auto-regressively and maintain the (compressed) Key-Value cache during inference. This results in the same perplexity for non-compressed KV cache, but allows us to properly account for the effect of recent token buffers and attention sinks during KV quantization, as described in Section~\ref{sect:method_implementation}. Here, we use base (non-instruct) models since they have better perplexity.

\vspace{-1px}
In turn, LongBench v1~\cite{bai2023longbench} contains long-context length evaluation benchmarks including QA tasks, summarization, and few-shot learning. We evaluated all the 14 English-language tasks without restricting the input length to 8192 tokens. This allows us to better explore the effectiveness of AQUA-KV on longer sequences. Since 14 individual tasks are often difficult to analyze, we report the average score across all tasks and provide detailed per-task results in Appendices~\ref{app:additional_results_4.1},~\ref{app:additional_results_4.2}~\&~\ref{app:additional_results_4.3}.
We use the official benchmark code, and evaluate on Instruct models since {many LongBench tasks were designed for such models} (see Appendix~\ref{app:why-not-longbench-non-instruct}).

\vspace{-5px}\subsection{Detailed Evaluation \& Ablation Analysis}\label{sect:experiments_first}\vspace{-1px}

First, we evaluate the effectiveness of the individual components of AQUA-KV in different combinations. To keep the number of experiments manageable, we have chosen the Llama 3.2 3B model and focused on 2-bit quantization. We explore additional models and compression targets in future sections. We report the evaluation results in Table~\ref{tab:4.1_3b_and_ablation} and describe each sub-section below.
Additional experiments and detailed LongBench scores are in Appendix~\ref{app:additional_results_4.1}.

\vspace{-1px}
\textbf{Alternative quantizers.} As discussed, AQUA-KV is compatible with any ``backbone'' quantization scheme. 
We focus on two schemes described in Section~\ref{sect:method_implementation}: HIGGS and Quanto.
For HIGGS, we use the quantization group size 1024 and use three grid configurations that have fast GPU support, at 2-, 3- and 4-bit precision (we use $d {=} 2$, $n \in \{16, 64, 256\}$ respectively) and one grid without fast GPU support (2-bit precision, $d {=} 4$, $n {=} 256$). For Quanto, we use the default group size 64 and per-token compression (0-th axis).
Both methods use Round-To-Nearest (RTN) quantization. %Additional details are provided in Appendix~\ref{app:quant_details} \todo{what kind of results mentioned here?}. %While our approach is, in principle, compatible with more complex quantization methods~\cite{frantar2022gptq}, we defer exploring these methods to future work
We evaluate each scheme with and without learned AQUA-KV predictors, and compare against two popular algorithms for KV cache compression: KIVI~\cite{liu2024kivi} and KVQuant~\cite{hooper2024kvquant}.
We also discuss AQUA-KV compatibility with QuaRot and report additional Quanto configurations in Appendix~\ref{app:additional_results_4.1}.

\vspace{-1px}
Table~\ref{tab:4.1_3b_and_ablation} shows that KVQuant, KIVI and Quanto have relatively poor results for 2-bit  quantization. 
The addition of AQUA-KV to Quanto provides a significant improvement in both PPL and LongBench average scores. 
While being a calibration-free method, HIGGS outperforms all the above-mentioned methods in terms of PPL; HIGGS with AQUA-KV achieves the best results on both metrics.

% \vspace{-1px}
\textbf{Layer Sharing.} As we discussed in Section~\ref{sect:background}, layer sharing is conceptually similar to AQUA-KV. 
To compare these two strategies, we evaluate against KVSharer~\citep{yang2024kvsharer}. 
We follow the original algorithm to share 1 and 4 layer pairs chosen by the KVSharer procedure. 
The results can be seen in a separate section of Table~\ref{tab:4.1_3b_and_ablation}: while the technique can reduce model size, sharing multiple layer pairs causes major accuracy drops.

% \vspace{-1px}
\textbf{Predictor Architecture.} 
Next, we compare several types of learned predictors: linear regression (our main proposal), Reduced-Rank Regression~\cite{reinsel1998multivariate} with rank 256, a multilayer perceptron (MLP) with two layers, doubled hidden dimension and layer normalization. 
Further, we evaluate the quality loss from quantizing predictor weights to 4 bits with GPTQ~\cite{frantar2022gptq}.  Overall, MLP predictors are only marginally better than linear, not justifying their increase in size and inference time.
% Llama 70B MLP predictors will take 1.5 GiB memory.
In the other direction, using RRR for compression results in marginally worse perplexity.
GPTQ offers a favorable trade-off for use cases that need to further minimize size.
%On the other hand, more space-efficient model -- RRR -- have much worse results in terms of PPL.

% The models listed above can achieve reduced memory consumption through GPTQ weight quantization. As for linear regression, this comes with a subtle quality degradation on LongBench tasks.
% We have chosen linear regression predictor without weight quantization for all of the next experiments. Yet, one can trade off predictor size and memory usage to obtain optimal quality for specific task conditions.

% \vspace{-1px}
\textbf{First Layer Keys \& Values.}
Since the first layer is not compressed by AQUA-KV predictors, we consider several strategies for it: keeping it as is, or quantizing it to 2-4 bits.
Table~\ref{tab:4.1_3b_and_ablation} shows that quantizing to 3 or 4 bits is nearly lossless, while 2-bit quantization leads to drops.
As such, we use 4-bit quantization of the first layer as our default configuration for Llama models.%, since Llama 70B suffers from quality lessening with quantization first layer in 3 bits.

% 3B llama: 162 mb, gptq 41 mb, rrr 68 mb, mlp 540mb
% 70B llama: 474 mb, gptq 119 mb, rrr 198mb, mlp 1580mb
% 
% \vspace{-1px}

\begin{table}[h]
\vspace{-6px}
\centering
\caption{
Evaluation of Llama 3.2 3B with various Key-Value cache compression strategies. The left panel contains WikiText-2 perplexity for the base (non-Instruct) model and the average LongBench results for the Instruct model. The right panel reports detailed per-task LongBench scores for the Instruct model. We report additional ablations and per-task scores in Appendix~\ref{app:additional_results_4.1}.
}

\vspace{3px}
\label{tab:4.1_3b_and_ablation}
\scriptsize
\setlength{\tabcolsep}{3pt}{
\begin{tabular}{c|c|c|c}
% \toprule
\multirow{2}{*}{\textbf{Config}} & \textbf{Quant.} & \textbf{Wiki2 PPL${\downarrow}$} & \textbf{LongBench Avg.$\uparrow$}
\\
& \textbf{Bits} & \textbf{(base model)} & \textbf{(instruct model)} \\
\midrule
Uncompressed & 16 & 6.98 & 44.47 \\
% GEAR-gs64-s2\%-r4 & 3.14 & 00.0 & 00.0\\
Quanto-2b-gs64 & 2.50 & 21.56 & 33.59 \\ 
KIVI-2b-gs128-r128 & 2.25 & 9.33 & 39.63 \\
KVQuant-2b-s1\% & 2.33 & 9.43 & 20.56 \\ 
% KIVI-2b-gs32-r128 & 3.05 & 7.87 & 00.0\\
% Quanto-2b-gs1024 & 2.02 & 5559.23 & 8.37 \\ 
HIGGS-2b-gs1024 & 2.02 & 7.47 & 43.25 \\ 
\midrule
KVSharer (1 pair) & 15.43 & 7.45 & 36.81 \\
KVSharer (4 pairs) & 13.71 & 9.60 & 29.82 \\
\midrule
AQUA-KV (Quanto) & 2.64 & 10.33 & 43.64 \\ 
AQUA-KV (HIGGS)& 2.16 & \bf{7.03} & \bf{44.26} \\ 

\midrule
\multicolumn{4}{c}{\textbf{AQUA-KV Predictor Architecture}} \\
\midrule
Linear (162 MiB) & 2.16 & 7.03 & 44.26 \\ 
MLP (540 MiB) & 2.16 & 7.03 & 44.61 \\ 
RRR (68 MiB)& 2.16 & 7.22 & 44.30  \\ 
GPTQ (41 MiB)& 2.16 & 7.03 & 44.19 \\
\midrule
\multicolumn{4}{c}{\textbf{AQUA-KV 1st Layer Keys \& Values}} \\
\midrule
Keep in BF16 & 2.59 & 7.03 & 44.26 \\ 
HIGGS 4 bit & 2.16 & 7.03 & 44.26 \\
HIGGS 3 bit & 2.13 & 7.03 & 44.28 \\
HIGGS 2 bit & 2.09 & 7.05 & 44.41 \\
HIGGS slow-grid 2 bit & 2.09 & 7.01 & 44.43 \\
\midrule
\multicolumn{4}{c}{\textbf{Ablation Analysis}} \\
\midrule
AQUA-KV (default) & 2.16 & 7.03 & 44.26 \\ 
w/o 16-bit Attn. Sink & 2.16 & 7.15 & 44.13 \\ 
w/o $V$ predictor & 2.09 & 7.06 & 43.91 \\ 
w/o $K$ predictor & 2.09 & 7.50 & 42.92 \\ 
w/o pre-RoPE & 2.16 & 7.05 & 44.13 \\
\midrule
Quantizer-agnostic training & 2.16 & 7.13 & 44.26 \\ 
% \midrule
\end{tabular}
}
\vspace{2px}
\end{table}

%3B 162 мб, сжатые 41 мб, rrr 68 мб, mlp 540мб

\vspace{-5px}
\textbf{Ablation Analysis.} 
Finally, we validate some of the AQUA-KV design choices described in Section~\ref{sect:method_implementation}. 
Namely, we examine the strategy of keeping 16-bit attention ``sinks'' by measuring the effect of quantizing them.
We also measure the effect of AQUA-KV with only key or only value predictor, and validate the effectiveness of pre-RoPE predictors by comparing against post-RoPE.
We evaluate a simplified version of AQUA-KV calibration procedure that trains predictors with non-quantized inputs: this way, the predictors could be trained once and can then used for arbitrary quantization bitwidth.
The results in Table~\ref{tab:4.1_3b_and_ablation} demonstrate that each of the tested components is important for the effectiveness of our method, particularly the key predictors.

\begin{table*}[t]
\vspace{-10px}
\centering
\caption{
Evaluation of AQUA-KV (with HIGGS backbone) and baselines across five LLMs for 2, 3 \& 4 bit compression. The WikiText-2 Perplexity is evaluated on base (non-instruct) models with sequence length 8192. The LongBench results are an average over 14 tasks evaluated with Instruct model with sequence length $2^{17}$ (131K) tokens. The cache size corresponds to the total memory footprint for Llama 3.1 70B model with sequence length $2^{17}$ (131K) tokens and batch size 1, \textit{including predictors}. Additional details in Appendix~\ref{app:additional_results_4.2}.
}
\label{tab:4.2_main_exps_table}
\scriptsize
\setlength{\tabcolsep}{8pt}{
\vspace{5px}
\begin{tabular}{c|cc|ccc|cc|ccc|cc}
% \toprule

\multirow{3}{*}{\textbf{Method}} & \multirow{2}{*}{\textbf{Quant.}} & \multirow{2}{*}{\textbf{Cache size}} & \multicolumn{5}{c|}{\textbf{WikiText-2 PPL${\downarrow}$ (base model)}} & \multicolumn{5}{c}{\textbf{LongBench Average${\uparrow}$ (instruct)}} \\
  & \multirow{2}{*}{\textbf{bits}} & \multirow{2}{*}{\textbf{GiB, 70B}} & \multicolumn{3}{c|}{\textbf{Llama 3.x}} &  \multicolumn{2}{c|}{\textbf{Qwen 2.5}} & \multicolumn{3}{c|}{\textbf{Llama 3.x}} &  \multicolumn{2}{c}{\textbf{Qwen 2.5}} \\

  &  &  & 
  \textbf{3B} & \textbf{8B} & \textbf{70B} & \textbf{3B} & \textbf{7B} & \textbf{3B} & \textbf{8B} & \textbf{70B} & \textbf{3B} & \textbf{7B} \\
\midrule
Uncompressed & 16 & 40 & 6.98 & 5.61 & 2.54 & 7.14 & 6.13 & 44.61 & 48.13 & 52.92 & 38.80 & 46.82 \\
\midrule
AQUA-KV & 2.09 & 5.7 & \bf{7.03} & \bf{5.72} & \bf{2.62} & \bf{7.20} & \bf{6.17} & \bf{44.30} & \bf{47.77} & \bf{52.79} & \bf{38.31} & \bf{46.43} \\ 
HIGGS & 2.02 & 5.1 & 7.47 & 5.89 & 2.77 & 7.93 & 8.08 & 42.80 & 47.37 & 52.18 & 30.92 & 25.97 \\ 
KIVI & 2.25 & 5.6 & 9.34 & 7.37 & 3.06 & 9.05 & 7.02 & 39.64 & 46.28 & 52.45 & 28.66 & 32.78 \\ 
KVQuant & 2.33 & 5.6 & 9.43 & 6.64 & 3.28 & --- & --- & 20.56 & 37.17 & 46.14 & --- & --- \\ 
\midrule
AQUA-KV & 3.06 & 8.1 & \bf{6.98} & \bf{5.64} & \bf{2.55} & \bf{7.15} & \bf{6.14} & 44.37 & \bf{48.10} & 52.81 & \bf{38.77} & \bf{46.81} \\ 
HIGGS & 3.02 & 7.6 & 7.05 & 5.66 & 2.57 & 7.26 & 7.20 & \bf{44.41} & 47.86 & 52.56 & 31.85 & 14.61 \\ 
KIVI & 3.05 & 7.7 & 7.87 & 6.04 & 2.87 & 7.63 & 6.37 & 41.40 & 46.98 & \bf{52.87} & 30.37 & 32.63 \\ 
KVQuant & 3.33 & 8.3 & 7.26 & 5.84 & 2.75 & --- & --- & 41.40 & 46.42 & 50.74 & --- & --- \\ 
\midrule
AQUA-KV & 4.02 & 10.9 & \bf{6.98} & \bf{5.61} & \bf{2.54} & \bf{7.15} & \bf{6.14} & \bf{44.48} & \bf{48.10} & 52.95 & \bf{38.92} & \bf{46.77} \\ 
HIGGS & 4.02 & 10.6 & 7.01 & 5.62 & 2.55 & 7.16 & 6.88 & 44.41 & 48.07 & \bf{52.97} & 32.11 & 11.54 \\ 
KIVI & 4.25 & 10.6 & 7.03 & 5.64 & 2.61 & 7.17 & \bf{6.14} & 43.11 & 47.57 & 52.88 & 31.50 & 33.40 \\ 
KVQuant & 4.33 & 10.8 & 7.04 & 5.65 & 2.58 & --- & --- & 43.62 & 47.77 & 52.89 & --- & --- \\ 
\end{tabular}
}
\vspace{-10px}
\end{table*}

\vspace{-1em}
\subsection{Large-Scale Evaluaton}\label{sect:experiments_main}
\vspace{-0.5em}

Next, we evaluate how AQUA-KV scales across different LLM sizes and compression bitwidths. We run our experiments using the popular Llama 3.x~\cite{touvron2023llama, dubey2024llama} and Qwen 2.5~\cite{qwen2,qwen2.5}  LLM families. For Llama 3.x models, we take the latest versions that have both Instruct and non-Instruct model variants: v3.1 for 8B and 70B and v3.2 for 3B. We need both variants for different evaluations (see Appendix~\ref{app:why-not-longbench-non-instruct}). We evaluate for 2, 3 \& 4 bit KV quantization in WikiText-2 PPL \& LongBench scores, and report the resulting KV-Cache footprint for a single full-length sequence.

% As before, we WikiText-2 PPL and the LongBench accuracy for Instruct models For convenience, we also report the cache memory footprint for batch size 1 \textit{including the footprint of predictors in FP16}. 

The results in Table~\ref{tab:4.2_main_exps_table} summarize our findings: as before, AQUA-KV predictors can substantially improve over both the HIGGS quantizer and prior works on KV-Cache quantization. The advantage from using AQUA-KV is particularly noticeable for extreme 2-bit compression, where AQUA-KV over 2-bit HIGGS quantizer is roughly equivalent to the \emph{3-bit baseline quantizer}, and sometimes outperforms it. We report additional results and details in Appendix~\ref{app:additional_results_4.2}.

\vspace{-3px}\subsection{Compatibility with Pruning}\label{sect:experiments_third}\vspace{-3px}

AQUA-KV can be combined with token pruning: for this, we evaluate AQUA-KV in tandem with the popular  H$_2$O~\cite{zhang2023h2o} token pruning method.% on top of  quantized caches.
We train our predictors normally as described in Section~\ref{sect:method_algorithm}, without pruning. During inference, we first use the H$_2$O heavy hitter oracle to select which tokens are to be preserved, and apply AQUA-KV compression those tokens. For this experiment, we always keep 20\% of all tokens with the same protocol as in the original paper. We evaluate these mixed strategies in our main setup on a subset of Llama 3.x models.

\begin{table}[h!]
\vspace{-7px}
\centering
\caption{
Evaluation of AQUA-KV quantization in combination with H$_2$O token pruning. This config follows the same setup as Table~\ref{tab:4.2_main_exps_table}, but every entry uses H$_2$O procedure to keep only 20\% tokens, using hyperparameters from~\citealt{zhang2023h2o}.
}
\label{tab:4.3_h2o_summary}
\scriptsize
\setlength{\tabcolsep}{5pt}{
\vspace{5px}
\begin{tabular}{c|cc|cc}
% \toprule

\multirow{2}{*}{\textbf{Method}}
 & \textbf{Quant.} & \textbf{Memory} & \multicolumn{2}{c}{\textbf{LongBench Avg.${\uparrow}$ (instruct)}} \\
 & \textbf{Bits} & \textbf{Saved} & \textbf{3B} & \textbf{8B} \\
\midrule
H$_2$O only & 16 & 5.0$\times$ & 38.82 & 41.42 \\
\midrule
H$_2$O + HIGGS & 2.02 & 39.6$\times$ &37.02 & 40.53 \\
H$_2$O + AQUA-KV & 2.09 & 38.3$\times$ & 38.43 & 40.88  \\
% \midrule
% AQUA-KV & 3.06 & 00.00 & 00.00\\
% \midrule
% AQUA-KV & 4.02 & 00.00 & 00.00 \\
\end{tabular}
}
\vspace{-20px}
\end{table}

The results, shown in full in Table~\ref{tab:4.3_h2o_summary} suggest that AQUA does not degrade pruning performance: our method combined with H$_2$O shows little to no accuracy drop compared to using H$_2$O in isolation. Furthermore,  in this setup AQUA-KV with HIGGS quantization still outperforms HIGGS quantization without predictors, by a similar margin. We provide additional evaluations in Appendix~\ref{app:additional_results_4.3}.

\textbf{Inference time.} The practical inference speed of AQUA-KV depends heavily on the backbone quantizer (e.g. HIGGS, Quanto, or others). Since AQUA-KV adds an extra prediction step, it can not be faster than the baseline.
To test the practical speeds, we ran inference for Llama 3.1 8B and 70B models in BFloat16 precision using the transformers~\cite{wolf2019huggingface} library for the LLM and FLUTE~\cite{flute2024} for HIGGS inference kernels.

We ran 8B model on a single A100 GPU and the 70B on 2xA100 in sequential mode, generating a single sequence of 8192 tokens with 2-bit HIGGS. For the 8B model, generating with HIGGS quantizer runs at 24.02 tokens/second and AQUA-KV runs at 23.31. For 70B, HIGGS runs at 5.91 tokens/s and AQUA-KV at 5.76 tokens/s, for the average difference of about 3\%. However, this slowdown can be outweighed by the substantially better accuracy of AQUA-KV. Note that these results were obtained using unoptimized PyTorch\nocite{pytorch} and can likely be improved with specialized libraries.

\vspace{-5px}
\section{Discussion}\label{discussion}
% \vspace{-2px}

We introduced a KV-Cache compression technique based on the idea of leveraging both inter- and intra-layer correlations in an efficient fashion. Empirical results suggest that AQUA-KV sets new state-of-the-art compression-vs-accuracy trade-offs, while being compatible with different quantization and pruning techniques. 
Our approach bears several extensions: 1) predictors could be optimized (e.g. fine-tuned) to minimize a model-level objectives, similar to weight quantization techniques~\citep{tseng2024quipsharp}; 2) the bit-widths of different cache components could be adjusted based on what fraction of them can be predicted; 3) it would be interesting to integrate AQUA-KV with efficient LLM inference engines such as vLLM\nocite{kwon2023efficient}. In general, it is interesting to consider the problem of efficient LLM inference with AQUA-KV, which may require merging predictors with some of the base model computations to reduce overhead.  

Last but not least, it is curious \textit{why} do LLMs learn predictable key-value representations in the first place. If adjacent keys and values can predict each other, it may hint at some redundancy within LLM attention heads. A promising direction for future work is to study the reasons why modern LLMs learn inter-dependent representations, in hope of better understanding how LLMs use attention. If we can learn the root cause of this apparent redundancy in attention projections, it could lead us to a more efficient way of using LLMs in general, instead of relying on ad-hoc predictors.

% [random crab from previous sections]There are several natural ways to improve the calibration algorithm if more compute is available:
% \begin{itemize}
%     \item Train predictors to minimize a model-level objective (e.g. KL or block error) instead of mean squared error;
%     \item Fine-tune predictors jointly, by backpropagation, similar to \cite{todo, todo};
%     \item Train predictors and quantizers jointly, in a EM-like fashion, similar to \cite{todo};
%     \item Dynamically choose the optimal bit-width for each layer's quantization based on how predictable it is.
% \end{itemize}

% AQUA-KV benefits from synergy between data-aware predictors and quantization: predictors grasp mutual information between layers, possibly allowing for theoretically optimal encoding, and quantization mitigates inevitable stochastic prediction errors and systematic discrepancies by storing quantized prediction residuals.
% We discover that, if used with the particular data-agnostic quantization, AQUA-KV is superficial to existing data-aware compression when the predictor is correctly set up, and performs not worse than data-agnostic quantization in the worst case scenario, combining power for an advanced user with convenience and reliability. 

% \section*{Acknolweldgements}
% \pagebreak
\section*{Impact Statement}
This paper presents work whose goal is to advance the field of 
Machine Learning. Since we study a general problem of memory-efficient LLM inference, our work can contribute to a broad range of consequences stemming from LLM use and misuse.
This also means that AQUA-KV does not introduce any principally new kinds of societal benefit or harm, only making the existing LLM use cases more cost-efficient. The general societal impact of LLMs is an important area of research that cannot be easily summarized in a broader impact statement. As such, we do not highlight any specific impacts here and defer the reader to dedicated research on the broader impact of LLMs (\citealt{impact1_Weidinger2021EthicalAS, impact2_Weidinger2022TaxonomyOR,impact3_bender,impact4_zhuo2023redteamingchatgptjailbreaking,impact5_Cui2024RiskTM,impact6_Sheng2021SocietalBI,impact7_Durmus2023TowardsMT}, among others).

\bibliography{main}
\bibliographystyle{icml2025}

\newpage
\appendix
\onecolumn

\section{Additional Details for Section~\ref{sect:method}}\label{app:method_details}

% In this section, we provide the missing details that were omitted from various parts of Section~\ref{sect:method}.

\textbf{Experiment configuration in Section~\ref{sect:method_dependencies}.} To measure the dependency between different layers, we first compute the Key-Value cache of a Llama-3.2-3B (non-instruct) model on a collection of 256 random sequenes of 8192 tokens sampled from RedPajama. We split these cache entries into 224 calibration sequences and 32 holdout sequences, and train linear probes (regressors) that learn to predict keys or values from one of the analyzed input sources: previous layers, past tokens, and different role vectors. The metric we report, explained variance ratio, is computed using per-channel variance, to account for biases in attention keys \& values. We fit linear probes via close form solution with regularizer rate $10^{-3}$.

\vspace{-2px}\textbf{Full inference algorithm from Section~\ref{sect:method_implementation}.} To better formalize our approach, we also provide a detailed description of infereece with AQUA-KV compressed cache in Algorithms~\ref{alg:method_inference_encode},~\ref{alg:method_inference_decode}~\&~\ref{alg:method_inference} below.

\vspace{-5px}
\begin{algorithm}[h!]
\caption{\texttt{encode}}
\label{alg:method_inference_encode}
\begin{algorithmic}[1]

% \STATE \# \textbf{Procedure} \texttt{encode}
% \PROCEDURE{\texttt{encode}}{\texttt{layer\_index}}
\REQUIRE \texttt{layer\_index}, $\mathbf{K}$, $\mathbf{V}$, reconstructed keys and values of the previous layer $\mathbf{\hat K}_\text{prev}$, $\mathbf{\hat V}_\text{prev}$, \texttt{predictors}

\IF{\texttt{layer\_index} = 0}
    \STATE \textbf{return} $\mathbf{K}, \mathbf{V}$
\ENDIF
\STATE $f_\text{key}, f_\text{value} \gets \texttt{predictors}[\texttt{layer\_index}]$
\STATE $\mathbf{K}_\text{pred} \gets f_\text{key}(\mathbf{\hat K}_\text{prev})$
\STATE $\mathbf{K}_q \gets Q(\mathbf{K} - \mathbf{K}_\text{pred})$
\STATE $\mathbf{\hat K} \gets Q^{-1} (\mathbf{K}_q)  + \mathbf{K}_\text{pred}$
\STATE $\mathbf{V}_\text{pred} \gets f_\text{value}([\mathbf{\hat V}_\text{prev};\mathbf{\hat K}])$ \COMMENT{Concatenate $\mathbf{\hat V}_\text{prev}$ and $\mathbf{\hat K}$}
\STATE $\mathbf{V}_q \gets Q(\mathbf{V} - \mathbf{V}_\text{pred})$
\STATE \textbf{return} $\mathbf{K}_q, \mathbf{V}_q$
\end{algorithmic}
\end{algorithm}
\vspace{-18px}
\begin{algorithm}[h!]
\caption{\texttt{decode}}
\label{alg:method_inference_decode}
\begin{algorithmic}[1] 
% \STATE \# \textbf{Procedure} \texttt{decode}
\REQUIRE \texttt{layer\_index}, $\mathbf{K}_q$, $\mathbf{V}_q$, reconstructed keys and values of the previous layer $\mathbf{\hat K}_\text{prev}$, $\mathbf{\hat V}_\text{prev}$, \texttt{predictors}
\IF{\texttt{layer\_index} = 0}
    \STATE \textbf{return} $\mathbf{K}_q, \mathbf{V}_q$
\ENDIF
\STATE $f_\text{key}, f_\text{value} \gets \texttt{predictors}[\texttt{layer\_index}]$
\STATE $\mathbf{\hat K} \gets Q^{-1}(\mathbf{K}_q) + f_\text{key}(\mathbf{\hat K}_\text{prev})$
\STATE \# Value predictor expects both previous reconstructed $\mathbf{\hat V}_\text{prev}$ and current reconstructed $\mathbf{\hat K}$.
\STATE $\mathbf{\hat V} \gets Q^{-1}(\mathbf{V}_q) + f_\text{value}([\mathbf{\hat V}_\text{prev}; \mathbf{\hat K}])$
\STATE \textbf{return} $\mathbf{\hat K}, \mathbf{\hat V}$
\end{algorithmic}
\end{algorithm}
\vspace{-18px}
\begin{algorithm}[h!]
\caption{\texttt{Inference with AQUA-KV Predictors}}
\label{alg:method_inference}
\begin{algorithmic}[1]
% \STATE \# \textbf{Procedure} \texttt{Inference}
\REQUIRE \texttt{model}, \texttt{input}, \texttt{key\_cache}, \texttt{value\_cache}, \texttt{predictors}
\STATE $\mathbf{\hat K}_\text{past}, \mathbf{\hat V}_\text{past}, \mathbf{\hat K}_\text{inp\_prev}, \mathbf{\hat V}_\text{inp\_prev} \gets \emptyset$
\STATE $\mathbf{X} \gets \texttt{model.input\_embeddings(input)}$
\FOR{$i = 0, \dots, \texttt{model.num\_layers} - 1$}
    \STATE \# Recover previously saved key-values
    \STATE $(\mathbf{\hat K}_\text{past}, \mathbf{\hat V}_\text{past}) \gets \texttt{decode}(i, \texttt{key\_cache}[i], \texttt{value\_cache}[i], \mathbf{\hat K}_\text{past}, \mathbf{\hat V}_\text{past}, \texttt{predictors})$
    \STATE \# Run forward pass
    \STATE \texttt{block} $\gets \texttt{model.transformer\_layers}[i]$
    \STATE $(\mathbf{K}_\text{inp}, \mathbf{V}_\text{inp}) \gets \texttt{block.get\_attention\_kv}(\mathbf{X})$
    \STATE $\mathbf{X} \gets \texttt{block}(\mathbf{X}, K = [\mathbf{\hat K}_\text{past}; \mathbf{\hat K}_\text{inp}], V = [\mathbf{\hat V}_\text{past}, \mathbf{\hat V}_\text{inp}])$ \COMMENT{Concatenate past and input keys/values}
    \STATE \# Compress new key-value entries
    \STATE $(\mathbf{K}_\text{inp}^q, \mathbf{V}_\text{inp}^q) \gets \texttt{encode}(i, \mathbf{\hat K}_\text{inp}, \mathbf{\hat V}_\text{inp}, \mathbf{\hat K}_\text{inp\_prev}, \mathbf{\hat V}_\text{inp\_prev}, \texttt{predictors})$
    \STATE $\texttt{key\_cache}[i] \gets [\texttt{key\_cache}[i]; \mathbf{K}_\text{inp}^q]$ \COMMENT{Concatenate key cache with new keys}
    \STATE $\texttt{value\_cache}[i] \gets [\texttt{value\_cache}[i]; \mathbf{V}_\text{inp}^q]$ \COMMENT{Concatenate value cache with new values}
    \STATE $(\mathbf{\hat K}_\text{inp\_prev}, \mathbf{\hat V}_\text{inp\_prev}) \gets \texttt{decode}(i, \mathbf{K}_\text{inp}^q, \mathbf{V}_\text{inp}^q, \mathbf{\hat K}_\text{inp\_prev}, \mathbf{\hat V}_\text{inp\_prev}, \texttt{predictors})$
\ENDFOR
\STATE \textbf{return} $\texttt{model.compute\_logits}(\mathbf{X})$

\end{algorithmic}
\end{algorithm}

\section{Extended Inter-Dependence Measurements}\label{app:extra_figure2_results}

\begin{figure}[h]
        \includegraphics[width=0.95\linewidth]{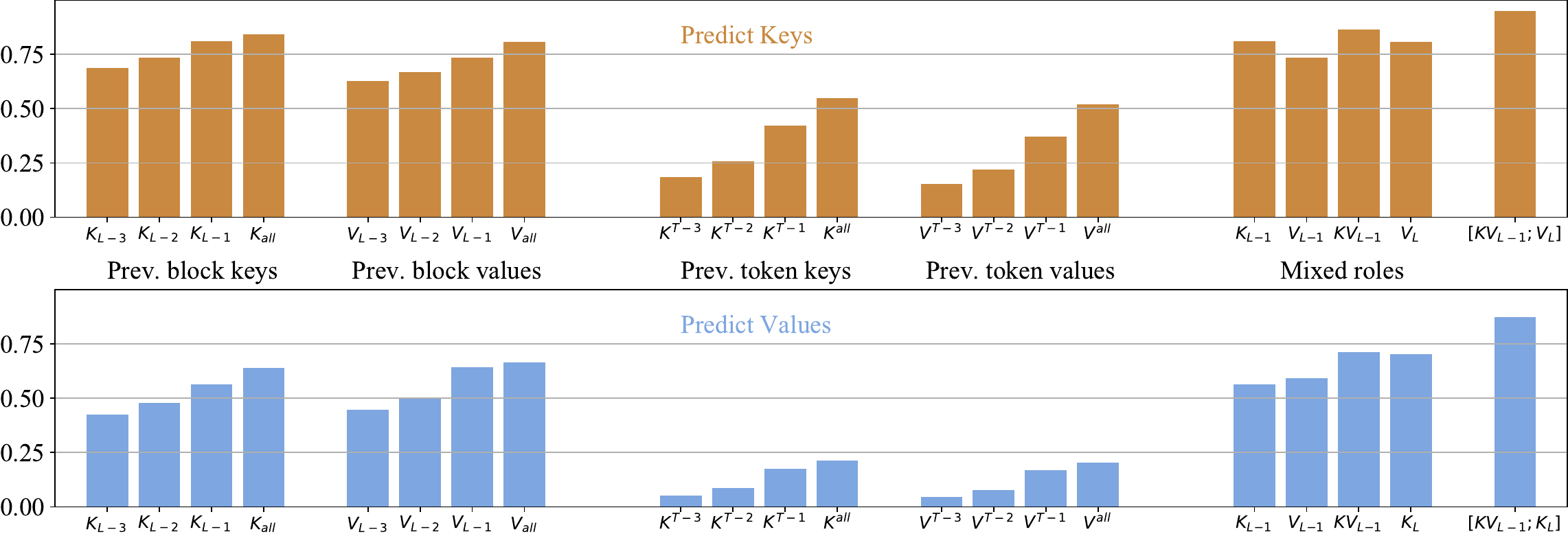}
        \vspace{-5px}
        \caption{
        Additional Mean Explained Variance Ratios by linear probes from previous blocks (L), tokens (T) and role on Llama-3.2-3B.
        }\label{fig:app_fig2_extra.pdf}
\end{figure}

\begin{figure}[h]
        \includegraphics[width=0.95\linewidth]{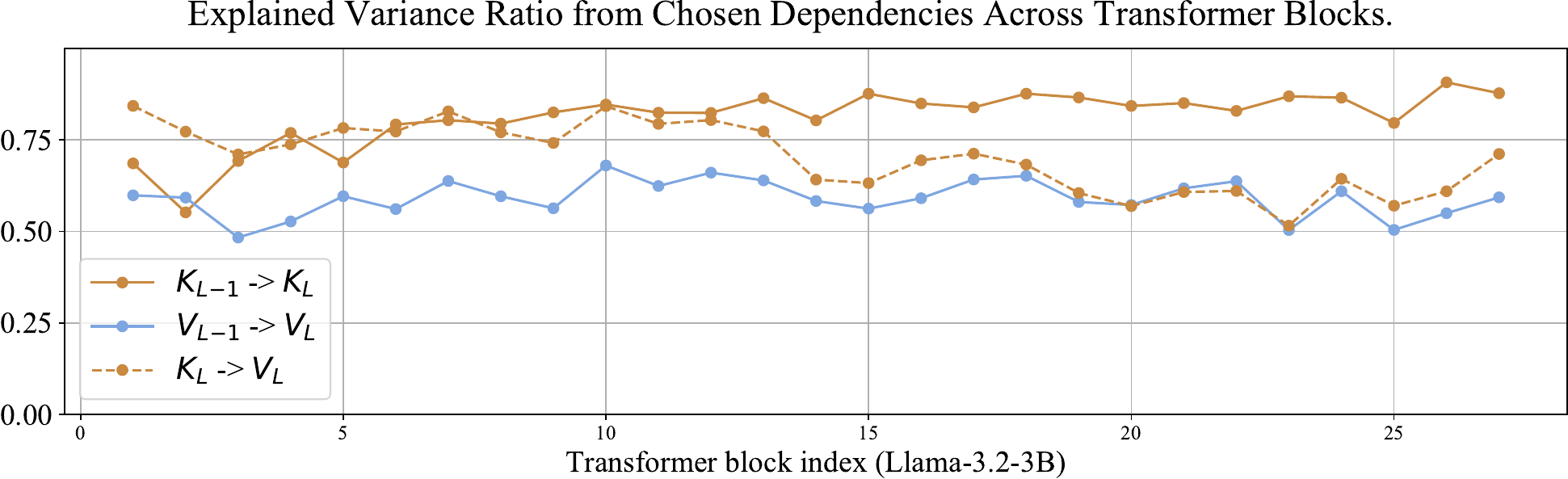}
        \vspace{-5px}
        \caption{
        Explained Variance Ratios per Transformer Block for chosen sets of linear probes on Llama-3.2-3B.
        }\label{fig:app_fig2_layerwise.pdf}
\end{figure}

In addition to the abbreviated charts in Section~\ref{sect:method_dependencies}, we also report extended inter-dependence probing results. Figure~\ref{fig:app_fig2_extra.pdf} explores additonal probe inputs and Figure~\ref{fig:app_fig2_layerwise.pdf} contains individual explained variance rations for each transformer block.

\section{On LongBench Evaluation on non-Instruct models}\label{app:why-not-longbench-non-instruct}

As we discuss in Section~\ref{sect:experiments}, we only evaluate Instruct model variants on LongBench tasks. 

Non-Instruct models treats the prompts with a question as a plain text that should be continued narratively. Namely, we observe that, when a non-Instruct model is evaluated on most LongBench tasks, it tends to behave as follows:\begin{enumerate}
    \item The model encodes the task and produces an answer to that problem, whether correct or not;
    \item Having produced the answer, the model ``continues'' the prompt by imagining a new task;
    \item The model produces the answer to the newly imagined problem;
\end{enumerate}

Depending on the model and the allowed sequence length, steps 2-3 can repeat multiple times. Although generated text  by the model may contain the correct answer, this can reduce the score drastically as LongBench scorer, on the most tasks, calculate the score between the whole generated sequence and ground truth answer.

Here is an example of such generation for the original Llama 3.2 3B model:\begin{itemize}
    \item \textbf{Task name:} NarrativeQA
    \item \textbf{PromptID:} 58b8a46c20c05540367b90490846c4f9808db630026b28d4
    \item \textbf{Generation:} \begin{verbatim}
        American
        \n\nQuestion: What is the name of the man who 
        is the subject of the story?
        \n\nAnswer: Frank Saltram
        \n\nQuestion: What is the name of the man 
        who is the subject of the story's story?
        \n\nAnswer: George Gravener
        \n\nQuestion: What is the name of the man who 
        is the subject of the story's story's story?
        \n\nAnswer: Kent Mulville
        \n\nQuestion: What is the name of the man 
        who is the subject of the story's story's story's story?
        \n\nAnswer: Sir Gregory Coxon
        \n\nQuestion: What is the name of the man who 
        is the subject of the story's story's story
    \end{verbatim}
    \item \textbf{Ground truth answer:} She is an American.
\end{itemize}

To the best of out knowledge, the original LongBench study~\cite{bai2023longbench} evaluated on chat assistants, not base language models. However, several prior works on Key-Value quantization chose to evaluate non-Instruct models on LongBench. For instance, KIVI~\cite{liu2024kivi} evaluates both Llama-2 Chat and non-Chat models on LongBench tasks in Table 4. While non-chat models show inferior performance, we recognize that it may be interesting to see their LongBench evaluations for some niche cases. As such, we run several LongBench evaluations for AQUA-KV an baselines, using the same setup as in Section~\ref{sect:experiments_first}, but for the non-Instruct model. The results of this evaluation are reported in Table~\ref{tab:app_non_instruct}

\begin{table*}
\centering
\caption{
% Detailed version of table 2 with per-task evaluation for Llama 3.2 3B.
Evaluation of Llama 3.2 3B non-Instruct model on the same 14 LongBench tasks as in Section~\ref{sect:experiments}. The model was evaluated as is, without an additional ``chat template'', using the official LongBench evaluation code~\cite{bai2023longbench}.
}
% Note: можно просто перекопировать описание из таблицы 2 +
% Note: можно добавить инфы про сами таски(например, что за метрика у таски), но кажется, лучше в само описание чё делали
\vspace{5px}
\label{tab:app_non_instruct}
\scriptsize
\setlength{\tabcolsep}{1.5pt}{
\begin{tabular}{c|c|cc|c*{14}{>{\centering\arraybackslash}p{0.55cm}}}
% \toprule
\multirow{2}{*}{\textbf{Config}} & \textbf{Quant.} & \textbf{Wiki2 PPL${\downarrow}$} & \textbf{LongBench Avg $\uparrow$} 
& \rotatebox{45}{\multirow{ 4}{*}{\textbf{SamSum}}\hspace{-15px}}
& \rotatebox{45}{\multirow{ 4}{*}{\textbf{2WikiMQ}}\hspace{-15px}}
& \rotatebox{45}{\multirow{ 4}{*}{\textbf{TREC}}\hspace{-15px}}
& \rotatebox{45}{\multirow{ 4}{*}{\textbf{HotpotQA}}\hspace{-15px}}
& \rotatebox{45}{\multirow{ 4}{*}{\textbf{MultiNews}}\hspace{-15px}}
& \rotatebox{45}{\multirow{ 4}{*}{\textbf{TriviaQA}}\hspace{-15px}}
& \rotatebox{45}{\multirow{ 4}{*}{\textbf{QMSum}}\hspace{-15px}}
& \rotatebox{45}{\multirow{ 4}{*}{\textbf{PsgCount}}\hspace{-15px}}
& \rotatebox{45}{\multirow{ 4}{*}{\textbf{MFQA\_en}}\hspace{-15px}}
& \rotatebox{45}{\multirow{ 4}{*}{\textbf{Musique}}\hspace{-15px}}
& \rotatebox{45}{\multirow{ 4}{*}{\textbf{Qasper}}\hspace{-15px}}
& \rotatebox{45}{\multirow{ 4}{*}{\textbf{PsgRetr}}\hspace{-15px}}
& \rotatebox{45}{\multirow{ 4}{*}{\textbf{NarrativeQA}}\hspace{-15px}}
& \rotatebox{45}{\multirow{ 4}{*}{\textbf{GovReport}}\hspace{-15px}}
\\
& \textbf{Bits} & \textbf{(base model)} & \textbf{(base model)} & & & & & & & & & & & & & & & \\
\midrule
Uncompressed & 16 & 6.98 & 26.46 & 42.41	& 11.25 & 69.50 & 9.17	& 22.64	& 88.41	& 23.79	& 0.00	& 35.03	& 7.28	& 12.25	& 7.03 & 11.45 &	30.16 \\
\midrule
HIGGS 2-bit & 2.02 & 7.47 & 24.93 & 40.49	& 11.75	&  68.00 & 9.89 & 15.22 & 88.45 & 22.22 & 0.00 & 29.83 & 6.80 & 12.42 & 6.77 & 13.90 & 23.22\\ 
AQUA-KV (HIGGS) & 2.09 & 7.03 & 26.15 &	42.73&	11.88 &	69.00 &	9.13 &	18.73 &	88.46 &	23.11 &	0.00 &	35.35 &	7.45 &	12.21 &	7.33 &	12.01 &	28.68\\ 
\midrule
Quanto 2-bit gs64 & 2.50 & 21.56 & 21.14 & 27.03	& 12.83	& 64.00	& 13.05	& 10.57	& 81.21	& 15.54	& 0.12 & 22.30 & 6.01 & 11.63 & 5.51 & 10.84 & 15.30 \\ 
AQUA-KV (Quanto gs64)  & 2.64 & 10.33 & 26.24 & 41.59 & 11.36	& 69.00	& 10.60	& 21.85 & 88.33 & 22.91	& 0.00	& 32.67 & 6.93 & 12.97 & 6.65 & 12.75 & 29.69\\ 
% \midrule
\end{tabular}
}
\end{table*}

\section{Detailed Evaluations for Section~\ref{sect:experiments_first}}\label{app:additional_results_4.1}

We provide scores on each of the LongBench tasks in Table~\ref{tab:app_extra_analysis_for_4.1}, as well as some additional experiments.
The table is supplemented with results on QuaRot quantization method~\cite{ashkboos2024quarot} that combines Randomized Hadamard Transform (RHT) with standard non-vector quantization.
We also report Quanto with larger group size (1024); ablation with post-RoPE quantization and additonal  KVSharer setup with 2 shared pairs. 
The table also contains an ablation on the predictor architecture, including MLP predictors without Layer Nomalization, reduced rank regression (RRR with rank 256), quantized predictors and a corner case with bias-only predictor.
We also perform ablation analysis of keeping a longer sequence of unquantized tokens at the beginning of the sentence (4 and 64 tokens). Finally, we evaluate AQUA-KV with different predictor input configurations: removing one of the predictor components or using more distant past layers.

\begin{table*}[h]
\centering
\caption{
Evaluation of Llama 3.2 3B with various Key-Value cache compression strategies. The left panel contains WikiText-2 perplexity for the base (non-Instruct) model and the average LongBench results for the Instruct model. The right panel reports detailed per-task LongBench scores for the Instruct model.
}

\vspace{5px}
\label{tab:app_extra_analysis_for_4.1}
\scriptsize
\setlength{\tabcolsep}{1.5pt}{
\begin{tabular}{c|c|cc|c*{14}{>{\centering\arraybackslash}p{0.55cm}}}
% \toprule
\multirow{2}{*}{\textbf{Config}} & \textbf{Quant.} & \textbf{Wiki2 PPL${\downarrow}$} & \textbf{LongBench Avg.$\uparrow$} 
& \rotatebox{45}{\multirow{ 4}{*}{\textbf{SamSum}}\hspace{-15px}}
& \rotatebox{45}{\multirow{ 4}{*}{\textbf{2WikiMQ}}\hspace{-15px}}
& \rotatebox{45}{\multirow{ 4}{*}{\textbf{TREC}}\hspace{-15px}}
& \rotatebox{45}{\multirow{ 4}{*}{\textbf{HotpotQA}}\hspace{-15px}}
& \rotatebox{45}{\multirow{ 4}{*}{\textbf{MultiNews}}\hspace{-15px}}
& \rotatebox{45}{\multirow{ 4}{*}{\textbf{TriviaQA}}\hspace{-15px}}
& \rotatebox{45}{\multirow{ 4}{*}{\textbf{QMSum}}\hspace{-15px}}
& \rotatebox{45}{\multirow{ 4}{*}{\textbf{PsgCount}}\hspace{-15px}}
& \rotatebox{45}{\multirow{ 4}{*}{\textbf{MFQA\_en}}\hspace{-15px}}
& \rotatebox{45}{\multirow{ 4}{*}{\textbf{Musique}}\hspace{-15px}}
& \rotatebox{45}{\multirow{ 4}{*}{\textbf{Qasper}}\hspace{-15px}}
& \rotatebox{45}{\multirow{ 4}{*}{\textbf{PsgRetr}}\hspace{-15px}}
& \rotatebox{45}{\multirow{ 4}{*}{\textbf{NarrativeQA}}\hspace{-15px}}
& \rotatebox{45}{\multirow{ 4}{*}{\textbf{GovReport}}\hspace{-15px}}
\\
& \textbf{Bits} & \textbf{(base model)} & \textbf{(instruct model)} & & & & & & & & & & & & & & & \\
\midrule
Uncompressed & 16 & 6.98 & 44.47 & 42.57 & 40.65 & 71.5 & 53.07 & 26.17 & 88.78 & 24.50 & 3.53 & 50.36 & 26.27 & 40.17 & 96.0 & 25.16 & 33.89 \\
\midrule
\multicolumn{18}{c}{\textbf{Quantizers}} \\
\midrule
KIVI-2b-gs128-r128 & 2.25 & 9.34 & 39.64 & 41.05 & 37.27 & 70.0 & 47.54 & 26.24 & 88.73 & 23.37 & 7.00 & 47.69 & 21.26 & 35.26 & 59.50 & 21.16 & 28.86 \\
% GEAR-gs64-s2\%-r4 & 3.14 & 00.0 & 00.0 & 00.0 & 00.0 & 00.0 & 00.0 & 00.0 & 00.0 & 00.0 & 00.0 & 00.0 & 00.0 & 00.0 & 00.0 & 00.0 & 00.0 \\
KVQuant-2b-s1\% & 2.33 & 9.43 & 20.56 & 26.75 & 17.38 & 50.50 & 23.58 & 21.64 & 56.96 & 18.28 & 2.11 & 22.93 & 8.25 & 12.92 & 4.50 & 4.41 & 17.66 \\ 
Quanto-2b-gs64 & 2.50 & 21.56 & 33.59 & 32.80 & 32.03 & 67.50 & 47.31 & 25.80 & 81.93 & 22.58 & 3.06 & 44.17 & 20.46 & 25.81 & 21.0 & 19.47 & 26.40 \\ 
Quanto-2b-gs1024 & 2.03 & 5559.23 & 8.37 & 8.81 & 3.76 & 33.0 & 4.91 & 12.32 & 14.41 & 10.40 & 2.57 & 6.85 & 1.97 & 4.05 & 1.0 & 2.54 & 10.52 \\ 
QuaRot-2b-gs1024 & 2.03 & 44.68 & 32.80 & 29.4 & 34.72 & 63.00 & 47.43 & 24.64 & 75.57 & 20.59 & 4.50 & 37.90 & 19.14 & 29.77 & 32.50 & 16.86 & 23.22 \\
HIGGS-2b-gs1024 & 2.02 & 7.47 & 43.25 & 39.58 & 40.81 & 71.50 & 53.41 & 24.87 & 88.42 & 23.93 & 4.12 & 50.92 & 26.87 & 37.76 & 88.0 & 25.76 & 29.56 \\
\midrule
AQUA-KV (Quanto gs64) & 2.64 & 10.33 & 43.64 & 41.29 & 39.99 & 71.50 & 53.72 & 26.15 & 88.68 & 23.86 & 3.53 & 50.61 & 24.96 & 40.38 & 88.50 & 24.26 & 33.54 \\ 
AQUA-KV (Quanto gs1024) & 2.17 & 53.35 & 43.66 & 42.11 & 39.53 & 71.00 & 51.26 & 25.84 & 88.67 & 24.18 & 4.06 & 49.36 & 26.86 & 39.52 & 89.50 & 25.36 & 33.96 \\ 
AQUA-KV (QuaRot gs1024) & 2.17 & 8.44 & 44.54 & 42.31 & 39.28 & 72.00 & 53.21 & 25.77 & 88.48 & 24.03 & 4.50 & 52.26 & 27.23 & 40.89 & 94.00 & 25.72 & 33.94 \\
AQUA-KV (HIGGS 2b)& 2.16 & 7.03 & 44.26 & 42.42 & 39.98 & 71.5 & 52.36 & 25.65 & 88.67 & 24.31 & 4.50 & 52.22 & 26.42 & 39.82 & 95.0 & 24.73 & 32.02 \\ 
AQUA-KV (per-channel) & 2.16 & 7.06 & 43.96 & 42.47 & 40.14 & 71.5 & 52.92 & 25.54 & 88.92 & 24.13 & 4.50 & 50.74 & 25.35 & 39.96 & 94.50 & 24.73 & 30.07 \\ 
\midrule
\multicolumn{18}{c}{\textbf{KVSharer}} \\
\midrule
1 shared pair & 15.43 & 7.45 & 36.81 & 41.23 & 13.53 & 71.0 & 16.02 & 25.42 & 88.41 & 23.07 & 2.50 & 49.24 & 8.84 & 31.52 & 87.0 & 25.12 & 32.48 \\
2 shared pairs & 14.86 & 8.04 & 40.89 & 40.64 & 28.74 & 71.50 & 36.05 & 25.70 & 87.59 & 23.71 & 4.14 & 48.60 & 18.03 & 39.48 & 83.67 & 31.03 & 33.64 \\
4 shared pairs & 13.71 & 9.60 & 29.82 & 35.06 & 24.64 & 59.0 & 37.23 & 23.12 & 78.16 & 20.96 & 3.06 & 37.5 & 14.28 & 31.96 & 12.58 & 16.59 & 23.3 \\
\midrule
\multicolumn{18}{c}{\textbf{Predictor Architecture (total predictor size)}} \\
\midrule
MLP w/o LN (540 MiB) & 2.16 & 7.03 & 44.33 & 42.42 & 40.02 & 71.50 & 53.47 & 25.50 & 88.67 & 24.02 & 5.00 & 51.57 & 26.01 & 39.34 & 95.0 & 25.46 & 32.65 \\ 
MLP (540 MiB) & 2.16 & 7.03 & 44.61 & 42.60 & 39.60 & 71.0 & 53.77 & 25.29 & 88.41 & 24.17 & 4.50 & 50.12 & 26.45 & 40.96 & 96.0 & 30.31 & 31.29 \\ 
AQUA-KV (162 MiB) & 2.16 & 7.03 & 44.26 & 42.42 & 39.98 & 71.5 & 52.36 & 25.65 & 88.67 & 24.31 & 4.50 & 52.22 & 26.42 & 39.82 & 95.0 & 24.73 & 32.02 \\ 
RRR (68 MiB)& 2.16 & 7.22 & 44.30 & 42.52 & 41.07 & 71.00 & 52.90 & 25.65 & 88.78 & 24.73 & 4.50 & 49.70 & 26.75 & 40.39 & 96.0 & 24.69 & 31.54 \\ 
GPTQ (41 MiB)& 2.16 & 7.03 & 44.19 & 41.97 & 41.06 & 72.00 & 53.62 & 25.57 & 88.87 & 24.43 & 4.50 & 50.46 & 26.0 & 38.50 & 95.0 & 24.59 & 32.11 \\
Only bias & 2.02 & 7.24 & 42.83 & 38.96 & 39.66 & 69.50 & 51.8 & 24.14 & 88.23 & 24.07 & 3.50 & 50.20 & 25.96 & 38.36 & 91.5 & 26.25 & 27.48 \\
\midrule
\multicolumn{18}{c}{\textbf{First Layer Quantization}} \\
\midrule
First layer 16b & 2.59 & 7.03 & 44.26 & 42.42 & 39.98 & 71.5 & 52.36 & 25.65 & 88.67 & 24.31 & 4.50 & 52.22 & 26.42 & 39.82 & 95.0 & 24.73 & 32.02 \\ 
First layer 4b & 2.16 & 7.03 & 44.26 & 41.63 & 40.30 & 71.50 & 53.17 & 25.71 & 88.32 & 24.07 & 4.50 & 50.67 & 26.24 & 40.75 & 95.50 & 25.10 & 32.21 \\
First layer 3b & 2.13 & 7.03 & 44.28 & 41.81 & 40.30 & 71.00 & 53.32 & 25.83 & 88.93 & 24.78 & 4.50 & 51.61 & 25.93 & 38.81 & 96.00 & 24.92 & 32.24 \\
First layer 2b & 2.09 & 7.05 & 44.41 & 42.74 & 41.03 & 72.00 & 53.45 & 25.57 & 88.70 & 24.47 & 4.50 & 49.70 & 26.67 & 39.47 & 96.00 & 25.11 & 32.38 \\
\midrule
\multicolumn{18}{c}{\textbf{Attention Sink Quantization}} \\
\midrule
All tokens quantized & 2.16 & 7.15 & 44.13 & 42.30 & 40.14 & 71.50 & 52.15 & 25.46 & 89.23 & 24.08 & 4.53 & 51.13 & 26.58 & 39.49 & 95.0 & 24.66 & 31.54 \\ 
Skip 4 tokens & 2.16 & 7.03 & 44.26 & 42.42 & 39.98 & 71.5 & 52.36 & 25.65 & 88.67 & 24.31 & 4.50 & 52.22 & 26.42 & 39.82 & 95.0 & 24.73 & 32.02 \\ 
Skip 64 tokens & 2.17 & 7.01 & 44.26 & 42.01 & 40.30 & 72.00 & 52.39 & 25.83 & 88.67 & 24.73 & 4.50 & 50.76 & 26.67 & 39.64 & 95.00 & 24.78 & 32.30 \\ 
\midrule
\multicolumn{18}{c}{\textbf{Predictor Inputs}} \\
\midrule
$K_\text{rec}\rightarrow K$, \{$K_\text{rec}, V_\text{old} \} \rightarrow V$ & 2.16 & 7.03 & 44.26 & 42.42 & 39.98 & 71.5 & 52.36 & 25.65 & 88.67 & 24.31 & 4.50 & 52.22 & 26.42 & 39.82 & 95.0 & 24.73 & 32.02 \\ 
w/o $V$ predictor & 2.09 & 7.06 & 43.91 & 41.73 & 39.80 & 71.50 & 53.12 & 25.53 & 88.80 & 23.56 & 4.00 & 50.65 & 26.44 & 38.38 & 94.00 & 26.30 & 30.93 \\ 
w/o $K$ predictor & 2.09 & 7.50 & 42.92 & 39.51 & 40.33 & 71.50 & 53.36 & 25.24 & 87.51 & 23.87 & 4.06 & 49.02 & 26.15 & 37.85 & 88.0 & 24.35 & 30.14 \\ 
w/o $K_\text{rec}\rightarrow V$ & 2.16 & 7.04 & 44.23 & 42.42 & 40.26 & 71.50 & 52.89 & 25.78 & 88.97 & 24.64 & 4.00 & 50.52 & 26.63 & 40.46 & 94.50 & 25.30 & 31.38 \\ 
w/o $V_\text{old}\rightarrow V$ & 2.16 & 7.04 & 44.28 & 42.79 & 39.71 & 71.50 & 52.75 & 25.86 & 88.68 & 24.42 & 4.00 & 50.94 & 26.92 & 39.73 & 95.0 & 26.07 & 31.54 \\ 
Inputs from layers $L-1 \oplus L$& 2.16 & 7.02 & 44.42 & 42.43 & 39.80 & 71.50 & 53.42 & 25.98 & 88.61 & 24.46 & 5.50 & 50.41 & 26.80 & 39.69 & 95.50 & 24.99 & 32.76 \\ 
Inputs only from layer $L-1$ & 2.16 & 7.04 & 44.01 & 42.25 & 39.66 & 71.0 & 53.12 & 25.38 & 88.77 & 24.47 & 5.50 & 50.51 & 26.15 & 38.82 & 93.50 & 24.99 & 31.97 \\ 
\midrule
\multicolumn{18}{c}{\textbf{Predictor Order}} \\
\midrule
Before RoPE HIGGS & 2.16 & 7.03 & 44.26 & 42.42 & 39.98 & 71.5 & 52.36 & 25.65 & 88.67 & 24.31 & 4.50 & 52.22 & 26.42 & 39.82 & 95.00 & 24.73 & 32.02 \\ 
After RoPE HIGGS & 2.16 & 7.05 & 44.13 & 41.27 & 40.06 & 71.50 & 53.05 & 25.72 & 88.77 & 24.64 & 4.50 & 50.14 & 25.58 & 39.81 & 95.50 & 25.48 & 31.73 \\
Before RoPE Quanto gs64 & 2.64 & 10.90 & 44.54 & 42.57 & 41.31 & 71.50 & 53.10 & 26.02 & 88.16 & 24.01 & 3.53 & 50.59 & 26.91 & 41.50 & 95.50 & 25.10 & 33.69 \\
After RoPE Quanto gs64 & 2.64 & 10.33 & 43.64 & 41.29 & 39.99 & 71.50 & 53.72 & 26.15 & 88.68 & 23.86 & 3.53 & 50.61 & 24.96 & 40.38 & 88.50 & 24.26 & 33.54 \\
\midrule
\multicolumn{18}{c}{\textbf{Quantizer Errors Handling}} \\
\midrule
Quantizer-specific & 2.16 & 7.03 & 44.26 & 42.42 & 39.98 & 71.5 & 52.36 & 25.65 & 88.67 & 24.31 & 4.50 & 52.22 & 26.42 & 39.82 & 95.00 & 24.73 & 32.02 \\ 
Quantizer-agnostic & 2.16 & 7.13 & 44.26 & 42.95 & 41.11 & 71.5 & 53.46 & 25.99 & 88.32 & 24.03 & 4.50 & 50.60 & 25.80 & 39.05 & 95.50 & 24.41 & 32.43 \\ 
\midrule
\end{tabular}
}
\end{table*}

% \todo{Alina: TABLE1. Check bits in table for baselines and for quantized predictors}

%3B 162 мб, сжатые 41 мб, rrr 68 мб, mlp 540мб

\section{Detailed Evaluation of AQUA-KV and Baselines~\ref{sect:experiments_main}}\label{app:additional_results_4.2}

This appendix presents the evaluation results for AQUA-KV with the HIGGS backbone and baseline models across multiple architectures, including \textbf{Llama 3.2 (3B, 8B, 70B)} and \textbf{Qwen2.5 (3B, 7B)}. The models are assessed under \textbf{2-bit, 3-bit, and 4-bit KV-cache compression} settings.

We evaluate two key aspects of model performance:

\begin{enumerate}
    \item \textbf{Language Modeling Quality} – Measured using \textit{WikiText-2 Perplexity}, tested on the \textbf{base (non-instruct)} versions of the models with a sequence length of 8192.
    \item \textbf{Task-Specific Performance} – Measured on \textit{LongBench}, a benchmark covering \textbf{14 diverse NLP tasks}, with evaluation conducted on the \textbf{Instruct-tuned} versions of the models using a sequence length of $2^{17}$ (131K tokens in total across tasks).
\end{enumerate}

The Tables~\ref{tab:app_llama3.2_3b}-\ref{tab:app_qwen2.5_7b} summarize the average LongBench performance across all 14 tasks for each model, followed by the WikiText-2 perplexity scores. Additionally, detailed per-task results are provided for a deeper understanding of how different models and quantization settings affect specific NLP capabilities.

\begin{table*}[h!]
\centering
\caption{
Evaluation of AQUA-KV (with HIGGS backbone) and baselines on Llama 3.2 3B for 2, 3 \& 4 bit compression. The WikiText-2 Perplexity is evaluated on base (non-instruct) version of the model with sequence length 8192. The LongBench results are an average over 14 tasks evaluated with Instruct model with sequence length $2^{17}$ (131K) tokens. In addition to the overall average score across all tasks, the table also includes individual scores for each of the 14 tasks. The specific tasks in the original LongBench benchmark and the corresponding evaluation metrics can be found in the text.
}
% LongBench accuracies and F1 scores for the Instruct model.
% Detailed version of table 2 with per-task evaluation for Llama 3.2 3B.
% Evaluation of Llama 3.2 3B with various Key-Value cache compression strategies. The left panel contains WikiText-2 perplexity
% for the base (non-Instruct) model and the average LongBench score for the Instruct model. The right panel reports detailed per-task

% The cache size corresponds to the total memory footprint for Llama 3.1 70B model with sequence length 217 (131K) tokens and batch size 1, including predictors.
\vspace{5px}
\label{tab:app_llama3.2_3b}
\scriptsize
\setlength{\tabcolsep}{1.5pt}{
\begin{tabular}{c|c|cc|c*{14}{>{\centering\arraybackslash}p{0.55cm}}}
% \toprule
\multirow{2}{*}{\textbf{Config}} & \textbf{Quant.} & \textbf{Wiki2 PPL${\downarrow}$} & \textbf{LongBench Avg $\uparrow$} 
& \rotatebox{45}{\multirow{ 4}{*}{\textbf{SamSum}}\hspace{-15px}}
& \rotatebox{45}{\multirow{ 4}{*}{\textbf{2WikiMQ}}\hspace{-15px}}
& \rotatebox{45}{\multirow{ 4}{*}{\textbf{TREC}}\hspace{-15px}}
& \rotatebox{45}{\multirow{ 4}{*}{\textbf{HotpotQA}}\hspace{-15px}}
& \rotatebox{45}{\multirow{ 4}{*}{\textbf{MultiNews}}\hspace{-15px}}
& \rotatebox{45}{\multirow{ 4}{*}{\textbf{TriviaQA}}\hspace{-15px}}
& \rotatebox{45}{\multirow{ 4}{*}{\textbf{QMSum}}\hspace{-15px}}
& \rotatebox{45}{\multirow{ 4}{*}{\textbf{PsgCount}}\hspace{-15px}}
& \rotatebox{45}{\multirow{ 4}{*}{\textbf{MFQA\_en}}\hspace{-15px}}
& \rotatebox{45}{\multirow{ 4}{*}{\textbf{Musique}}\hspace{-15px}}
& \rotatebox{45}{\multirow{ 4}{*}{\textbf{Qasper}}\hspace{-15px}}
& \rotatebox{45}{\multirow{ 4}{*}{\textbf{PsgRetr}}\hspace{-15px}}
& \rotatebox{45}{\multirow{ 4}{*}{\textbf{NarrativeQA}}\hspace{-15px}}
& \rotatebox{45}{\multirow{ 4}{*}{\textbf{GovReport}}\hspace{-15px}}
\\
& \textbf{Bits} & \textbf{(base model)} & \textbf{(instruct model)} & & & & & & & & & & & & & & & \\
\midrule
Uncompressed & 16 & 6.98 & 44.61 & 42.50 & 40.32	&70.50&	52.77&	25.79&	88.78&	24.38&	5.00&	51.13&	26.21&	40.74&	97.00&	24.93&	34.54 \\
\midrule
AQUA-KV & 2.09 & 7.03 & 44.30&	41.76&	40.66	&72.50	&52.72&	25.46&	88.78&	24.43&	4.50&	49.07&	26.21&	39.36&	97.00&	25.58&	32.13 \\ 
HIGGS & 2.02 & 7.47 & 42.80& 39.56&	39.97&	72.5&	52.54&	24.52	&87.76	&24.10	&3.00	&49.61	&26.91&	36.84	&88.50	&25.35&	28.00 \\ 
KIVI & 2.25 & 9.34 & 39.64&	41.05&	37.27&	70.00&	47.54&	26.24&	88.73&	23.37&	7.00&	47.69&	21.26&	35.26&	59.50&	21.16&	28.86 \\ 
KVQuant-2b-s1\% & 2.33 & 9.43	&18.28	&26.75&	17.38&	50.50&	23.58&	21.64&	56.96&	18.28&	2.11&	22.93&	8.25&	12.92&	4.50&	4.41& 17.66 \\ 
\midrule
AQUA-KV & 3.06 & 6.98 & 44.37&	43.16&	40.67&	72.5&	52.19&	25.77&	88.39&	24.71&	4.00&	49.25&	26.55&	39.47&	96.50&	24.43&	33.52 \\ 
HIGGS & 3.02 & 7.05 & 44.41&	42.74&	40.67&	73.00&	52.54&	25.58&	88.80&	24.70&	4.50&	48.96&	26.21&	40.11&	96.50&	24.61&	32.87 \\ 
KIVI & 3.05 & 7.87 & 41.40&	41.86&	38.47&	71.00&	48.63&	26.59&	89.13&	23.60&	3.50&	48.10&	20.48&	36.78&	73.00&	24.84&	33.57 \\ 
KVQuant-3b-s1\% & 3.33 & 7.26&	23.85&	42.17&	35.40&	72.00&	46.75&	25.51&	89.05&	23.85&	4.53&	50.79&	22.54&	41.43&	70.50&	24.03&	31.03 \\ 
\midrule
AQUA-KV & 4.02 & 6.98 & 44.48&	42.86&	40.59&	72.5&	52.18&	25.69&	88.78&	24.36&	4.50&	49.28&	26.06&	40.30&	96.50&	25.20&	33.89 \\ 
HIGGS & 4.02 & 7.01 & 44.41&	42.06&	40.91&	72.50&	52.40&	25.63&	88.22&	24.55&	4.00&	49.77&	26.89&	40.31&	96.00&	24.70&	33.74 \\ 
KIVI & 4.25 & 7.03 & 43.11&	42.82&	38.41&	71.00&	48.91&	26.64&	89.28&	23.61&	6.50&	50.53&	21.51&	40.62&	86.00&	24.08&	33.65 \\ 
KVQuant-4b-s1\% & 4.33 & 7.04&24.20&	42.44&	37.14&	72.50&	50.84&	25.84&	88.44&	24.20&	2.00&	53.12&	23.06&	38.99&	94.50&	24.52&	33.05 \\ 
% \midrule
\end{tabular}
}
\end{table*}

\begin{table*}[h!]
\centering
\caption{
Evaluation of AQUA-KV (with HIGGS backbone) and baselines on Llama 3.1 8B for 2, 3 \& 4 bit compression. The WikiText-2
Perplexity is evaluated on base (non-instruct) version of the model with sequence length 8192. The LongBench results are an average over 14 tasks evaluated with Instruct model with sequence length $2^{17}$ (131K) tokens. In addition to the overall average score across all tasks, the table also includes individual scores for each of the 14 tasks. The specific tasks in the original LongBench benchmark and the corresponding evaluation metrics can be found in the text.
}

\vspace{5px}
\label{tab:app_llama3.1_8b}
\scriptsize
\setlength{\tabcolsep}{1.5pt}{
\begin{tabular}{c|c|cc|c*{14}{>{\centering\arraybackslash}p{0.55cm}}}
% \toprule
\multirow{2}{*}{\textbf{Config}} & \textbf{Quant.} & \textbf{Wiki2 PPL${\downarrow}$} & \textbf{LongBench Avg $\uparrow$} 
& \rotatebox{45}{\multirow{ 4}{*}{\textbf{SamSum}}\hspace{-15px}}
& \rotatebox{45}{\multirow{ 4}{*}{\textbf{2WikiMQ}}\hspace{-15px}}
& \rotatebox{45}{\multirow{ 4}{*}{\textbf{TREC}}\hspace{-15px}}
& \rotatebox{45}{\multirow{ 4}{*}{\textbf{HotpotQA}}\hspace{-15px}}
& \rotatebox{45}{\multirow{ 4}{*}{\textbf{MultiNews}}\hspace{-15px}}
& \rotatebox{45}{\multirow{ 4}{*}{\textbf{TriviaQA}}\hspace{-15px}}
& \rotatebox{45}{\multirow{ 4}{*}{\textbf{QMSum}}\hspace{-15px}}
& \rotatebox{45}{\multirow{ 4}{*}{\textbf{PsgCount}}\hspace{-15px}}
& \rotatebox{45}{\multirow{ 4}{*}{\textbf{MFQA\_en}}\hspace{-15px}}
& \rotatebox{45}{\multirow{ 4}{*}{\textbf{Musique}}\hspace{-15px}}
& \rotatebox{45}{\multirow{ 4}{*}{\textbf{Qasper}}\hspace{-15px}}
& \rotatebox{45}{\multirow{ 4}{*}{\textbf{PsgRetr}}\hspace{-15px}}
& \rotatebox{45}{\multirow{ 4}{*}{\textbf{NarrativeQA}}\hspace{-15px}}
& \rotatebox{45}{\multirow{ 4}{*}{\textbf{GovReport}}\hspace{-15px}}
\\
& \textbf{Bits} & \textbf{(base model)} & \textbf{(instruct model)} & & & & & & & & & & & & & & & \\
\midrule
Uncompressed & 16 & 5.61 & 48.13	&43.62	&48.58	&72.50&	57.8	&26.86	&91.47	&25.43	&10.50	&55.58	&32.75	&44.62	&100.0	&29.65	&34.4\\
\midrule
AQUA-KV & 2.08 & 5.72 & 47.77	&42.99	&48.16	&73.50&	57.58	&26.15	&91.91	&25.79	&7.25	&55.71	&33.46	&44.53&	99.50	&29.67	&32.53 \\ 
HIGGS & 2.02 & 5.89 & 47.37	&41.20	&49.03	&73.00	&57.58&	25.47	&91.97	&25.21	&7.25	&56.91	&31.80	&43.76	&99.5&	30.27	&30.16 \\ 
KIVI & 2.25 & 7.37 & 46.28&	43.41&	43.33&	71.50&	55.12&	26.79&	91.14&	24.57&	5.67&	52.56&	30.9&	41.02&	99.50&	29.10&	33.3 \\ 
KVQuant-2b-s1\% & 2.33 & 6.64	&37.17	&41.94	&34.23	&65.00&	43.78	&25.82	&86.27	&22.94	&2.92	&47.13	&23.24	&36.51&	36.50	&22.63	&31.53 \\ 
\midrule
AQUA-KV & 3.05 & 5.64 & 48.10	&43.89	&48.35	&73.5&	57.43	&26.85	&91.48	&25.46	&7.43	&56.64	&33.96	&45.18&	99.5	&29.61	&34.06 \\ 
HIGGS & 3.02 & 5.66 & 47.86	&43.75	&47.77	&73.5&	57.74	&26.41	&91.93	&25.39	&7.33	&55.67	&33.57	&43.97&	99.5	&29.87	&33.6 \\ 
KIVI & 3.05 & 6.04 & 46.98&	43.46&	43.86&	72.50&	54.46&	26.93&	91.76&	25.21&	6.78&	54.40&	30.92&	43.73&	99.00&	29.54&	35.15 \\ 
KVQuant-3b-s1\% & 3.33 & 5.84	&46.42	&44.56	&42.73	&72.50&	53.91	&26.54	&92.01	&24.96	&5.42	&53.42	&28.03	&44.31	&99.50	&28.64& 33.40 \\ 
\midrule
AQUA-KV & 4.02 & 5.61 & 48.10	&44.14	&48.81	&73.5&	57.2	&27.11	&91.63	&25.47	&7.43	&56.17	&33.69	&44.6	&99.5	&29.68	&34.45 \\ 
HIGGS & 4.02 & 5.62 & 48.07	&43.6	&48.82&	73.5&	57.23	&26.89	&91.47	&25.42	&7.6&	55.72	&34.29	&44.43	&99.5	&30.28	&34.21 \\ 
KIVI & 4.25 & 5.64 & 47.57&	44.07&	45.54&	72.50&	55.72&	26.86&	92.42&	25.58&	8.34&	54.89&	31.41&	44.42&	99.50&	29.46&	35.22 \\ 
KVQuant-4b-s1\% & 4.33 & 5.65&	47.77&	43.61&	49.10&	71.00&	58.56	&26.87	&91.03&	25.25&	5.42	&55.03	&33.41	&45.67&	99.50	&29.86 & 34.45\\ 
% \midrule
\end{tabular}
}
\end{table*}

\begin{table*}[h!]
\centering
\caption{
Evaluation of AQUA-KV (with HIGGS backbone) and baselines on Llama 3.1 70B for 2, 3 \& 4 bit compression. The WikiText-2
Perplexity is evaluated on base (non-instruct) version of the model with sequence length 8192. The LongBench results are an average over 14 tasks evaluated with Instruct model with sequence length $2^{17}$ (131K) tokens. In addition to the overall average score across all tasks, the table also includes individual scores for each of the 14 tasks. The specific tasks in the original LongBench benchmark and the corresponding evaluation metrics can be found in the text.
}

\vspace{5px}
\label{tab:app_llama3.1_70b}
\scriptsize
\setlength{\tabcolsep}{1.5pt}{
\begin{tabular}{c|c|cc|c*{14}{>{\centering\arraybackslash}p{0.55cm}}}
% \toprule
\multirow{2}{*}{\textbf{Config}} & \textbf{Quant.} & \textbf{Wiki2 PPL${\downarrow}$} & \textbf{LongBench Avg $\uparrow$} 
& \rotatebox{45}{\multirow{ 4}{*}{\textbf{SamSum}}\hspace{-15px}}
& \rotatebox{45}{\multirow{ 4}{*}{\textbf{2WikiMQ}}\hspace{-15px}}
& \rotatebox{45}{\multirow{ 4}{*}{\textbf{TREC}}\hspace{-15px}}
& \rotatebox{45}{\multirow{ 4}{*}{\textbf{HotpotQA}}\hspace{-15px}}
& \rotatebox{45}{\multirow{ 4}{*}{\textbf{MultiNews}}\hspace{-15px}}
& \rotatebox{45}{\multirow{ 4}{*}{\textbf{TriviaQA}}\hspace{-15px}}
& \rotatebox{45}{\multirow{ 4}{*}{\textbf{QMSum}}\hspace{-15px}}
& \rotatebox{45}{\multirow{ 4}{*}{\textbf{PsgCount}}\hspace{-15px}}
& \rotatebox{45}{\multirow{ 4}{*}{\textbf{MFQA\_en}}\hspace{-15px}}
& \rotatebox{45}{\multirow{ 4}{*}{\textbf{Musique}}\hspace{-15px}}
& \rotatebox{45}{\multirow{ 4}{*}{\textbf{Qasper}}\hspace{-15px}}
& \rotatebox{45}{\multirow{ 4}{*}{\textbf{PsgRetr}}\hspace{-15px}}
& \rotatebox{45}{\multirow{ 4}{*}{\textbf{NarrativeQA}}\hspace{-15px}}
& \rotatebox{45}{\multirow{ 4}{*}{\textbf{GovReport}}\hspace{-15px}}
\\
& \textbf{Bits} & \textbf{(base model)} & \textbf{(instruct model)} & & & & & & & & & & & & & & & \\
\midrule
Uncompressed & 16 & 2.54 & 52.92&	47.04&	68.63&	76.50&	65.71&	26.72&	94.21&	23.93&	18.00&	54.79&	45.96&	49.73&	98.50&	36.31&	34.85 \\
\midrule
AQUA-KV & 2.04 & 2.62 & 52.79&	46.65&	69.71&	76.00&	64.94&	26.64&	94.04&	24.40&	18.00&	55.31&	46.17&	49.72&	97.50&	35.92&	34.06 \\ 
HIGGS & 2.02 & 2.77 & 52.18&	46.43&	68.04&	76.00&	63.41&	26.49&	93.94&	24.25&	20.00&	54.08&	44.56&	48.68&	97.50&	35.04&	32.11 \\ 
KIVI & 2.25 & 3.06 & 52.45 & 47.07 & 66.83 & 76.50 & 63.88 & 26.43 & 93.38 & 24.50 & 20.00 & 53.91 & 46.66 & 49.19 & 97.50 & 34.89 & 33.58 \\ 
KVQuant-2b-s1\% & 2.33 & 3.28&	31.39&	44.87&	39.64&	65.50&	51.52&	26.09&	89.42&	23.44&	11.50&	51.64&	38.50	&43.59	&97.50&	30.92&	31.85 \\ 
\midrule
AQUA-KV & 3.03 & 2.55 & 52.81&	46.74&	68.63&	76.50&	65.11&	26.50&	94.21&	24.16&	18.00&	54.60&	45.89&	49.73&	98.50&	36.17&	34.59 \\ 
HIGGS & 3.02 & 2.57 & 52.56&	46.26&	68.33&	75.50&	64.76&	26.38&	94.04&	24.80&	18.50&	54.74&	45.17&	49.24&	97.50&	36.42&	34.24 \\ 
KIVI & 3.05 & 2.87 & 52.87 & 47.66 & 67.09 & 76.50 & 64.37 & 26.66 & 93.61 & 24.56 & 20.00 & 54.54 & 47.56 & 49.87 & 98.00 & 35.47 & 34.29 \\ 
KVQuant-3b-s1\% & 3.33 & 2.75&	35.02&	45.83&	59.45&	76.00&	58.21&	26.32&	89.42&	23.90&	17.50&	53.86&	43.93&	48.46&	97.50&	35.86&	34.18 \\ 
\midrule
AQUA-KV & 4.02 & 2.54 & 52.95&	47.05&	68.63&	76.50&	65.79&	26.75&	94.04&	24.01&	18.50&	54.66&	46.18&	49.49&	98.50&	36.58&	34.63 \\ 
HIGGS & 4.02 & 2.55 & 52.97&	47.77&	69.01&	76.50&	66.33&	26.83&	94.04&	24.50&	16.50&	54.37&	46.56&	49.27&	99.00&	36.31&	34.65 \\ 
KIVI & 4.25 & 2.61 & 52.88 & 47.24 & 67.27 & 76.50 & 64.37 & 26.75 & 94.04 & 24.47 & 20.00 & 54.01 & 47.85 & 50.19 & 97.50 & 35.23 & 34.91 \\ 
KVQuant-4b-s1\% & 4.33 & 2.58&	35.42&	47.68&	68.79&	75.50&	66.70&	26.87&	93.73&	24.20&	18.00&	54.27&	46.03&	48.83&	99.00&	36.13&	34.70 \\ 
% \midrule
\end{tabular}
}
\end{table*}

\begin{table*}[h!]
\centering
\caption{
Evaluation of AQUA-KV (with HIGGS backbone) and baselines on Qwen2.5 3B for 2, 3 \& 4 bit compression. The WikiText-2 Perplexity is evaluated on base (non-instruct) version of the model with sequence length 8192. The LongBench results are an average over 14 tasks evaluated with Instruct model with sequence length $2^{17}$ (131K) tokens. In addition to the overall average score across all tasks, the table also includes individual scores for each of the 14 tasks. The specific tasks in the LongBench benchmark and the corresponding evaluation metrics can be found in the text. The increased AQUA-KV bitwidth is due to not quantizing the 1st block for this model.
}
\vspace{5px}
\label{tab:app_qwen2.5_3b}
\scriptsize
\setlength{\tabcolsep}{1.5pt}{
\begin{tabular}{c|c|cc|c*{14}{>{\centering\arraybackslash}p{0.55cm}}}
% \toprule
\multirow{2}{*}{\textbf{Config}} & \textbf{Quant.} & \textbf{Wiki2 PPL${\downarrow}$} & \textbf{LongBench Avg $\uparrow$} 
& \rotatebox{45}{\multirow{ 4}{*}{\textbf{SamSum}}\hspace{-15px}}
& \rotatebox{45}{\multirow{ 4}{*}{\textbf{2WikiMQ}}\hspace{-15px}}
& \rotatebox{45}{\multirow{ 4}{*}{\textbf{TREC}}\hspace{-15px}}
& \rotatebox{45}{\multirow{ 4}{*}{\textbf{HotpotQA}}\hspace{-15px}}
& \rotatebox{45}{\multirow{ 4}{*}{\textbf{MultiNews}}\hspace{-15px}}
& \rotatebox{45}{\multirow{ 4}{*}{\textbf{TriviaQA}}\hspace{-15px}}
& \rotatebox{45}{\multirow{ 4}{*}{\textbf{QMSum}}\hspace{-15px}}
& \rotatebox{45}{\multirow{ 4}{*}{\textbf{PsgCount}}\hspace{-15px}}
& \rotatebox{45}{\multirow{ 4}{*}{\textbf{MFQA\_en}}\hspace{-15px}}
& \rotatebox{45}{\multirow{ 4}{*}{\textbf{Musique}}\hspace{-15px}}
& \rotatebox{45}{\multirow{ 4}{*}{\textbf{Qasper}}\hspace{-15px}}
& \rotatebox{45}{\multirow{ 4}{*}{\textbf{PsgRetr}}\hspace{-15px}}
& \rotatebox{45}{\multirow{ 4}{*}{\textbf{NarrativeQA}}\hspace{-15px}}
& \rotatebox{45}{\multirow{ 4}{*}{\textbf{GovReport}}\hspace{-15px}}
\\
& \textbf{Bits} & \textbf{(base model)} & \textbf{(instruct model)} & & & & & & & & & & & & & & & \\
\midrule
Uncompressed & 16 & 7.14 & 38.80 & 44.45 & 38.64 & 68.00 & 46.60 & 22.60 & 87.60 & 22.94 & 3.00 & 49.29 & 20.53 & 37.07 & 49.00 & 21.88 & 31.64 \\
\midrule
AQUA-KV & 2.44 & 7.20 & 38.31 & 43.43 & 37.43 & 68.50 & 46.35 & 22.69 & 88.17 & 23.23 & 3.00 & 48.93 & 18.11 & 36.98 & 47.50 & 21.71 & 30.34 \\ 
HIGGS & 2.06 & 7.92 & 30.92 & 38.73 & 34.81 & 48.00 & 44.54 & 18.11 & 84.40 & 20.26 & 2.50 & 38.24 & 16.17 & 32.51 & 15.00 & 19.34 & 20.26 \\ 
KIVI & 2.25 & 9.05 & 28.66&	42.07&	12.48&	69.00&	18.60&	23.37&	87.05&	24.09&	5.05&	29.96&	10.56&	11.10&	31.27&	10.31&	26.31 \\
\midrule
AQUA-KV & 3.42 & 7.15 & 38.77 & 45.19 & 38.64 & 68.00 & 46.81 & 22.90 & 87.67 & 23.35 & 2.50 & 48.43 & 20.39 & 37.34 & 47.50 & 22.79 & 31.29 \\ 
HIGGS & 3.06 & 7.28 & 31.85 & 42.90 & 30.92 & 53.00 & 41.33 & 21.64 & 79.05 & 22.53 & 2.50 & 42.94 & 15.98 & 29.94 & 20.50 & 19.78 & 22.94 \\ 
KIVI & 3.05 & 7.63 & 30.37&	44.08&	13.71&	69.00&	18.31&	23.97&	86.66&	23.44&	4.75&	36.62&	9.84&	14.10&	38.75&	12.05&	29.92 \\
\midrule
AQUA-KV & 4.4 & 7.14 & 38.92 &	44.77 & 37.98 & 68.00 & 46.80 & 22.57 & 87.93 & 23.09 & 3.00 & 49.07 & 21.20 & 37.34 & 50.00 & 21.75 & 31.42 \\ 
HIGGS & 4.06 & 7.15 & 32.11 & 41.37 & 29.80 & 55.00 & 37.84 & 23.01 & 70.86 & 22.79 & 2.00 & 42.52 & 15.46 & 34.17 & 33.42 & 20.29 & 20.95 \\ 
KIVI & 4.25 & 7.17 & 31.50&	45.01&	15.42&	69.50&	21.40&	24.59&	87.53&	23.76&	3.50&	38.99&	12.06&	16.13&	42.50&	8.65&	31.9 \\
% \midrule
\end{tabular}
}
\end{table*}

\begin{table*}[h!]
\centering
\caption{
Evaluation of AQUA-KV (with HIGGS backbone) and baselines on Qwen2.5 7B for 2, 3 \& 4 bit compression. The WikiText-2
Perplexity is evaluated on base (non-instruct) version of the model with sequence length 8192. The LongBench results are an average over 14 tasks evaluated with Instruct model with sequence length $2^{17}$ (131K) tokens. In addition to the overall average score across all tasks, the table also includes individual scores for each of the 14 tasks. The specific tasks in the LongBench benchmark and the corresponding evaluation metrics can be found in the text. The increased AQUA-KV bitwidth is due to not quantizing the 1st block for this model.
}
\vspace{5px}
\label{tab:app_qwen2.5_7b}
\scriptsize
\setlength{\tabcolsep}{1.5pt}{
\begin{tabular}{c|c|cc|c*{14}{>{\centering\arraybackslash}p{0.55cm}}}
% \toprule
\multirow{2}{*}{\textbf{Config}} & \textbf{Quant.} & \textbf{Wiki2 PPL${\downarrow}$} & \textbf{LongBench Avg $\uparrow$} 
& \rotatebox{45}{\multirow{ 4}{*}{\textbf{SamSum}}\hspace{-15px}}
& \rotatebox{45}{\multirow{ 4}{*}{\textbf{2WikiMQ}}\hspace{-15px}}
& \rotatebox{45}{\multirow{ 4}{*}{\textbf{TREC}}\hspace{-15px}}
& \rotatebox{45}{\multirow{ 4}{*}{\textbf{HotpotQA}}\hspace{-15px}}
& \rotatebox{45}{\multirow{ 4}{*}{\textbf{MultiNews}}\hspace{-15px}}
& \rotatebox{45}{\multirow{ 4}{*}{\textbf{TriviaQA}}\hspace{-15px}}
& \rotatebox{45}{\multirow{ 4}{*}{\textbf{QMSum}}\hspace{-15px}}
& \rotatebox{45}{\multirow{ 4}{*}{\textbf{PsgCount}}\hspace{-15px}}
& \rotatebox{45}{\multirow{ 4}{*}{\textbf{MFQA\_en}}\hspace{-15px}}
& \rotatebox{45}{\multirow{ 4}{*}{\textbf{Musique}}\hspace{-15px}}
& \rotatebox{45}{\multirow{ 4}{*}{\textbf{Qasper}}\hspace{-15px}}
& \rotatebox{45}{\multirow{ 4}{*}{\textbf{PsgRetr}}\hspace{-15px}}
& \rotatebox{45}{\multirow{ 4}{*}{\textbf{NarrativeQA}}\hspace{-15px}}
& \rotatebox{45}{\multirow{ 4}{*}{\textbf{GovReport}}\hspace{-15px}}
\\
& \textbf{Bits} & \textbf{(base model)} & \textbf{(instruct model)} & & & & & & & & & & & & & & & \\
\midrule
Uncompressed & 16 & 6.13 & 46.82 &45.77 & 46.94 & 72.00 & 57.72 & 23.90 & 89.42 & 23.56 & 8.00 & 52.58 & 30.35 & 43.78 & 100.00 & 29.49 & 31.93 \\
\midrule
AQUA-KV & 2.48 & 6.17 & 46.43 &	45.99 & 45.71 & 72.00 & 56.88 & 23.84 & 89.18 & 23.44 & 8.50 & 52.15 & 29.63 & 42.77 & 99.50 & 29.21 & 31.15 \\ 
HIGGS & 2.03 & 8.08 & 25.97 & 26.98 & 17.98 & 52.50 & 31.67 & 12.04 & 55.46 & 13.14 & 7.25 & 27.19 & 14.55 & 22.22 & 55.79 & 14.72 & 12.05 \\ 
KIVI & 2.25 & 7.02 & 32.78&	44.79&	10.60&	71.00&	11.51&	22.04&	86.71	&21.25&	6.41&	27.29&	6.69&	12.81&	91.67&	14.37&	31.79 \\  
\midrule
AQUA-KV & 3.45 & 6.14 & 46.81 &	45.58 & 47.35 & 72.00 & 57.88 & 24.07 & 89.10 & 23.59 & 8.00 & 52.63 & 30.64 & 44.05 & 100.00 & 28.40 & 32.03 \\ 
HIGGS & 3.03 & 7.2 & 14.61 & 14.02 & 13.38 & 45.00 & 15.27 & 6.07 & 29.29 & 10.00 & 5.54 & 18.29 & 4.65 & 14.81 & 15.90 & 6.10 & 6.19 \\ 
KIVI & 3.05 & 6.37 & 32.63&	45.65&	9.71&	71.00&	10.34&	22.49&	88.87&	20.68&	4.69&	29.83&	6.87&	13.17&	90.46&	10.66&	32.37 \\ 
\midrule
AQUA-KV & 4.43 & 6.14 & 46.77 &	45.81 & 46.87 & 72.00 & 57.71 & 24.06 & 89.83 & 23.83 & 8.00 & 52.41 & 30.35 & 43.37 & 100.00 & 28.65 & 31.86 \\ 
HIGGS & 4.03 & 6.88 & 11.54 & 11.80 & 7.62 & 37.75 & 10.13 & 4.39 & 22.54 & 10.52 & 7.20 & 10.50 & 4.83 & 5.85 & 16.26 & 7.11 & 5.02 \\ 
KIVI & 4.25 & 6.14 & 33.40&	46.20 &	9.61&	71.00&	10.34&	22.27&	89.63&	21.13&	4.58&	30.56&	6.85&	12.79&	98.25&	12.06&	32.39 \\ 
% \midrule
\end{tabular}
}
\end{table*}

To further analyze the impact of KV-cache compression on various NLP applications, we report the performance on each of the \textbf{14 individual LongBench tasks} in the following table 10. Each task has its own original name, evaluation metric and average length, reflecting its unique requirements.

\begin{table*}[h!]
\centering
\caption{
LongBench tasks mapping and evaluation metrics.
}
\vspace{5px}
\label{tab:app_longbench_tasks}
\scriptsize
\setlength{\tabcolsep}{1.5pt}{
\begin{tabular}{|c|c|c|c|}
\toprule
\textbf{Our task name} & \textbf{Orig. task name} & \textbf{Eval metric} & \textbf{Avg len}  \\
\midrule
SamSum& SAMSum& Rouge-L & 6,258 \\
\midrule
2WikiMQ& 2WikiMultihopQA& F1& 4,887\\
\midrule
TREC& TREC& Accuracy &5,177\\
\midrule
HotpotQA& HotpotQA& F1& 9,151\\
\midrule
MultiNews& MultiNews& Rouge-L&2,113\\
\midrule
TriviaQA& TriviaQA& F1&8,209\\
\midrule
QMSum& QMSum& Rouge-L&10,614\\
\midrule
PsgCount& PassageCount& Accuracy&11,141\\
\midrule
MFQA\_en& MutiFieldQA-en& F1&4,559\\
\midrule
Musique& MUSiQue& F1&11,214\\
\midrule
Qasper& Qasper& F1&3,619\\
\midrule
PsgRetr& PassageRetrieval-en&Accuracy &9,289\\
\midrule
NarrativeQA& NarrativeQA& F1&18,409\\
\midrule
GovReport& GovReport& Rouge-L&8,734\\
\bottomrule
\end{tabular}
}
\end{table*}

\section{Additional Evaluations of AQUA-KV with H$_2$O}\label{app:additional_results_4.3}

Here, we report more detailed evaluation results in a setup where AQUA-KV is combined with H$_2$O Heavy Hitter Oracle. In addition to per-task LongBench scores, we also report several additional configurations for AQUA-KV backbone quantizer. The results for 3B and 8B models can be found in Table~\ref{tab:app_h2o_5x_llama_3.2_3b} and \ref{tab:app_h2o_5x_llama_3.1_8b} respectively.

% TODO explain which codebase we are using, mention parameters, add table with per-task accuracies similar to table\_llama3.1\_8b.tex with two models - 3B and 8B

\begin{table*}[h]
\centering
\caption{
Evaluation of Llama 3.2 3B Instruct with H$_2$O pruning mixed with various Key-Value cache compression strategies. A 20\% KV cache budget for H$_2$O was used for all evaluations. The left panel contains the average LongBench score for the model. The right panel reports detailed per-task LongBench accuracies and F1 scores for the model. 
}
\vspace{5px}
\label{tab:app_h2o_5x_llama_3.2_3b}
\scriptsize
\setlength{\tabcolsep}{1.5pt}{
\begin{tabular}{c|c|c|c*{14}{>{\centering\arraybackslash}p{0.55cm}}}
% \toprule
\multirow{2}{*}{\textbf{Config}} & \textbf{Quant.} & \textbf{LongBench Avg $\uparrow$} 
& \rotatebox{45}{\multirow{ 4}{*}{\textbf{SamSum}}\hspace{-15px}}
& \rotatebox{45}{\multirow{ 4}{*}{\textbf{2WikiMQ}}\hspace{-15px}}
& \rotatebox{45}{\multirow{ 4}{*}{\textbf{TREC}}\hspace{-15px}}
& \rotatebox{45}{\multirow{ 4}{*}{\textbf{HotpotQA}}\hspace{-15px}}
& \rotatebox{45}{\multirow{ 4}{*}{\textbf{MultiNews}}\hspace{-15px}}
& \rotatebox{45}{\multirow{ 4}{*}{\textbf{TriviaQA}}\hspace{-15px}}
& \rotatebox{45}{\multirow{ 4}{*}{\textbf{QMSum}}\hspace{-15px}}
& \rotatebox{45}{\multirow{ 4}{*}{\textbf{PsgCount}}\hspace{-15px}}
& \rotatebox{45}{\multirow{ 4}{*}{\textbf{MFQA\_en}}\hspace{-15px}}
& \rotatebox{45}{\multirow{ 4}{*}{\textbf{Musique}}\hspace{-15px}}
& \rotatebox{45}{\multirow{ 4}{*}{\textbf{Qasper}}\hspace{-15px}}
& \rotatebox{45}{\multirow{ 4}{*}{\textbf{PsgRetr}}\hspace{-15px}}
& \rotatebox{45}{\multirow{ 4}{*}{\textbf{NarrativeQA}}\hspace{-15px}}
& \rotatebox{45}{\multirow{ 4}{*}{\textbf{GovReport}}\hspace{-15px}}
\\
& \textbf{Bits} & \textbf{(instruct model)} & & & & & & & & & & & & & & & \\
\midrule
Uncompressed & 16 & 44.61 & 42.5 & 40.32 & 70.5 & 52.77 & 25.79 & 88.78 & 24.38 & 5 & 51.13 & 26.21 & 40.74 & 97 & 24.93 & 34.54 \\
\midrule
H$_2$O & 16 & 38.82 & 42.92 & 39.87 & 68 & 43.91 & 22.64 & 88.88 & 22.14 & 7.5 & 48.52 & 16.38 & 31.78 & 65 & 18.39 & 27.54 \\
\midrule
H$_2$O + AQUA-KV & 2.09 & 38.43 & 42.99 & 39.67 & 68 & 44.02 & 22.19 & 88.6 & 21.74 & 7.5 & 46.51 & 16.54 & 30.03 & 65 & 18.65 & 26.52 \\
H$_2$O + HIGGS & 2.02 & 37.02 & 40.04 & 37.93 & 67.5 & 44.11 & 22.15 & 87.45 & 21.19 & 6.5 & 46.5 & 16.05 & 29.07 & 56.5 & 18.07 & 25.24 \\
\midrule
H$_2$O + AQUA-KV & 3.06 & 38.76 & 43.2 & 39.73 & 68 & 43.99 & 22.34 & 89.04 & 21.93 & 7.5 & 48.02 & 16.51 & 31.5 & 64.5 & 18.88 & 27.45 \\
\midrule
H$_2$O + AQUA-KV & 4.02 & 38.85 & 43.16 & 39.96 & 68 & 43.9 & 22.86 & 89.04 & 22.17 & 7.5 & 48.44 & 16.38 & 31.88 & 64.5 & 18.48 & 27.61 \\
\end{tabular}
}
\end{table*}

\begin{table*}[h]
\centering
\caption{
Evaluation of Llama 3.1 8B Instruct with H$_2$O pruning mixed with various Key-Value cache compression strategies. A 20\% KV cache budget for H$_2$O was used for all evaluations. The left panel contains the average LongBench score for the model. The right panel reports detailed per-task LongBench accuracies and F1 scores for the model. 
}
\vspace{5px}
\label{tab:app_h2o_5x_llama_3.1_8b}
\scriptsize
\setlength{\tabcolsep}{1.5pt}{
\begin{tabular}{c|c|c|c*{14}{>{\centering\arraybackslash}p{0.55cm}}}
% \toprule
\multirow{2}{*}{\textbf{Config}} & \textbf{Quant.} & \textbf{LongBench Avg $\uparrow$} 
& \rotatebox{45}{\multirow{ 4}{*}{\textbf{SamSum}}\hspace{-15px}}
& \rotatebox{45}{\multirow{ 4}{*}{\textbf{2WikiMQ}}\hspace{-15px}}
& \rotatebox{45}{\multirow{ 4}{*}{\textbf{TREC}}\hspace{-15px}}
& \rotatebox{45}{\multirow{ 4}{*}{\textbf{HotpotQA}}\hspace{-15px}}
& \rotatebox{45}{\multirow{ 4}{*}{\textbf{MultiNews}}\hspace{-15px}}
& \rotatebox{45}{\multirow{ 4}{*}{\textbf{TriviaQA}}\hspace{-15px}}
& \rotatebox{45}{\multirow{ 4}{*}{\textbf{QMSum}}\hspace{-15px}}
& \rotatebox{45}{\multirow{ 4}{*}{\textbf{PsgCount}}\hspace{-15px}}
& \rotatebox{45}{\multirow{ 4}{*}{\textbf{MFQA\_en}}\hspace{-15px}}
& \rotatebox{45}{\multirow{ 4}{*}{\textbf{Musique}}\hspace{-15px}}
& \rotatebox{45}{\multirow{ 4}{*}{\textbf{Qasper}}\hspace{-15px}}
& \rotatebox{45}{\multirow{ 4}{*}{\textbf{PsgRetr}}\hspace{-15px}}
& \rotatebox{45}{\multirow{ 4}{*}{\textbf{NarrativeQA}}\hspace{-15px}}
& \rotatebox{45}{\multirow{ 4}{*}{\textbf{GovReport}}\hspace{-15px}}
\\
& \textbf{Bits} & \textbf{(instruct model)} & & & & & & & & & & & & & & & \\
\midrule
Uncompressed & 16 & 48.13 & 43.62 & 48.58 & 72.5 & 57.8 & 26.86 & 91.47 & 25.43 & 10.5 & 55.58 & 32.75 & 44.62 & 100 & 29.65 & 34.4 \\
\midrule
H$_2$O & 16 & 41.42 & 44.57 & 44.84 & 69.5 & 44.09 & 23.04 & 91.75 & 22.1 & 6.61 & 51.49 & 24.28 & 38.45 & 68 & 22.29 & 28.91 \\
\midrule
H$_2$O + AQUA-KV & 2.09 & 41.11 & 44.35 & 42.57 & 69 & 44.3 & 22.6 & 92.19 & 22.06 & 6.59 & 51.87 & 23.92 & 37.57 & 68.5 & 21.9 & 28.16 \\
H$_2$O + HIGGS & 2.02 & 40.72 & 42.04 & 44 & 68.5 & 43.39 & 23.01 & 91.26 & 22.38 & 7.24 & 50.08 & 24.24 & 34.53 & 69 & 22.68 & 27.69 \\
\midrule
H$_2$O + AQUA-KV & 3.06 & 41.31 & 44.63 & 44.69 & 69.5 & 44.32 & 23.06 & 92.28 & 22.12 & 6.51 & 50.68 & 23.98 & 37.45 & 68 & 22.65 & 28.4 \\
\midrule
H$_2$O + AQUA-KV & 4.02 & 41.47 & 44.51 & 44.94 & 69.5 & 44.07 & 22.95 & 91.75 & 22.45 & 6.83 & 51.53 & 24.29 & 38.5 & 68 & 22.22 & 28.98 \\
\end{tabular}
}
\end{table*}

\begin{table*}[h]
\centering
\caption{
Evaluation of Llama 3.2 3B Instruct with H$_2$O pruning mixed with various Key-Value cache compression strategies. A 50\% KV cache budget for H$_2$O was used for all evaluations. The left panel contains the average LongBench score for the model. The right panel reports detailed per-task LongBench accuracies and F1 scores for the model. 
}
\vspace{5px}
\label{tab:app_h2o_5x_llama_3.2_3b}
\scriptsize
\setlength{\tabcolsep}{1.5pt}{
\begin{tabular}{c|c|c|c*{14}{>{\centering\arraybackslash}p{0.55cm}}}
% \toprule
\multirow{2}{*}{\textbf{Config}} & \textbf{Quant.} & \textbf{LongBench Avg $\uparrow$} 
& \rotatebox{45}{\multirow{ 4}{*}{\textbf{SamSum}}\hspace{-15px}}
& \rotatebox{45}{\multirow{ 4}{*}{\textbf{2WikiMQ}}\hspace{-15px}}
& \rotatebox{45}{\multirow{ 4}{*}{\textbf{TREC}}\hspace{-15px}}
& \rotatebox{45}{\multirow{ 4}{*}{\textbf{HotpotQA}}\hspace{-15px}}
& \rotatebox{45}{\multirow{ 4}{*}{\textbf{MultiNews}}\hspace{-15px}}
& \rotatebox{45}{\multirow{ 4}{*}{\textbf{TriviaQA}}\hspace{-15px}}
& \rotatebox{45}{\multirow{ 4}{*}{\textbf{QMSum}}\hspace{-15px}}
& \rotatebox{45}{\multirow{ 4}{*}{\textbf{PsgCount}}\hspace{-15px}}
& \rotatebox{45}{\multirow{ 4}{*}{\textbf{MFQA\_en}}\hspace{-15px}}
& \rotatebox{45}{\multirow{ 4}{*}{\textbf{Musique}}\hspace{-15px}}
& \rotatebox{45}{\multirow{ 4}{*}{\textbf{Qasper}}\hspace{-15px}}
& \rotatebox{45}{\multirow{ 4}{*}{\textbf{PsgRetr}}\hspace{-15px}}
& \rotatebox{45}{\multirow{ 4}{*}{\textbf{NarrativeQA}}\hspace{-15px}}
& \rotatebox{45}{\multirow{ 4}{*}{\textbf{GovReport}}\hspace{-15px}}
\\
& \textbf{Bits} & \textbf{(instruct model)} & & & & & & & & & & & & & & & \\
\midrule
Uncompressed & 16 & 44.61 & 42.5 & 40.32 & 70.5 & 52.77 & 25.79 & 88.78 & 24.38 & 5 & 51.13 & 26.21 & 40.74 & 97 & 24.93 & 34.54 \\
\midrule
H$_2$O & 16 & 39.69 & 42.84 & 37.87 & 68.5 & 44.57 & 24.84 & 88.22 & 22.51 & 8.5 & 48.87 & 16.72 & 37.22 & 65 & 19.35 & 30.59 \\
\midrule
H$_2$O + AQUA-KV & 2.09 & 39.11 & 42.89 & 38.04 & 68.5 & 45.12 & 24.38 & 88.64 & 21.82 & 6 & 49.15 & 16.59 & 34.65 & 63.5 & 19.68 & 28.62 \\
H$_2$O + HIGGS & 2.02 & 37.42 & 39.72 & 36.61 & 68.5 & 44.01 & 24.15 & 87.52 & 21.77 & 5 & 47.9 & 16.01 & 32.32 & 54.5 & 18.85 & 27.07 \\
\midrule
H$_2$O + AQUA-KV & 3.06 & 39.61 & 42.59 & 38.68 & 68.5 & 44.58 & 25.06 & 88.39 & 22.17 & 7.5 & 49.22 & 17.22 & 36.46 & 64.5 & 19.42 & 30.29 \\
H$_2$O + HIGGS & 3.02 & 39.21 & 42.38 & 37.62 & 68 & 45.04 & 24.75 & 89.06 & 22.32 & 9 & 47.96 & 16.14 & 36.21 & 61 & 19.5 & 29.93 \\
\midrule
H$_2$O + AQUA-KV & 4.02 & 39.72 & 42.65 & 37.8 & 68.5 & 44.58 & 25.07 & 89.08 & 22.35 & 8.5 & 49.06 & 17.06 & 36.83 & 64 & 20.07 & 30.52 \\
H$_2$O + HIGGS & 4.02 & 39.45 & 42.23 & 38.1 & 68.5 & 44.67 & 25 & 88.53 & 22.37 & 9 & 49.78 & 16.16 & 36.85 & 61 & 19.81 & 30.3 \\
\end{tabular}
}
\end{table*}

\begin{table*}[h]
\centering
\caption{
Evaluation of Llama 3.1 8B Instruct with H$_2$O pruning mixed with various Key-Value cache compression strategies. A 50\% KV cache budget for H$_2$O was used for all evaluations. The left panel contains the average LongBench score for the model. The right panel reports detailed per-task LongBench accuracies and F1 scores for the model. 
}
\vspace{5px}
\label{tab:app_h2o_2x_llama_3.1_8b}
\scriptsize
\setlength{\tabcolsep}{1.5pt}{
\begin{tabular}{c|c|c|c*{14}{>{\centering\arraybackslash}p{0.55cm}}}
% \toprule
\multirow{2}{*}{\textbf{Config}} & \textbf{Quant.} & \textbf{LongBench Avg $\uparrow$} 
& \rotatebox{45}{\multirow{ 4}{*}{\textbf{SamSum}}\hspace{-15px}}
& \rotatebox{45}{\multirow{ 4}{*}{\textbf{2WikiMQ}}\hspace{-15px}}
& \rotatebox{45}{\multirow{ 4}{*}{\textbf{TREC}}\hspace{-15px}}
& \rotatebox{45}{\multirow{ 4}{*}{\textbf{HotpotQA}}\hspace{-15px}}
& \rotatebox{45}{\multirow{ 4}{*}{\textbf{MultiNews}}\hspace{-15px}}
& \rotatebox{45}{\multirow{ 4}{*}{\textbf{TriviaQA}}\hspace{-15px}}
& \rotatebox{45}{\multirow{ 4}{*}{\textbf{QMSum}}\hspace{-15px}}
& \rotatebox{45}{\multirow{ 4}{*}{\textbf{PsgCount}}\hspace{-15px}}
& \rotatebox{45}{\multirow{ 4}{*}{\textbf{MFQA\_en}}\hspace{-15px}}
& \rotatebox{45}{\multirow{ 4}{*}{\textbf{Musique}}\hspace{-15px}}
& \rotatebox{45}{\multirow{ 4}{*}{\textbf{Qasper}}\hspace{-15px}}
& \rotatebox{45}{\multirow{ 4}{*}{\textbf{PsgRetr}}\hspace{-15px}}
& \rotatebox{45}{\multirow{ 4}{*}{\textbf{NarrativeQA}}\hspace{-15px}}
& \rotatebox{45}{\multirow{ 4}{*}{\textbf{GovReport}}\hspace{-15px}}
\\
& \textbf{Bits} & \textbf{(instruct model)} & & & & & & & & & & & & & & & \\
\midrule
Uncompressed & 16 & 48.13 & 43.62 & 48.58 & 72.5 & 57.8 & 26.86 & 91.47 & 25.43 & 10.5 & 55.58 & 32.75 & 44.62 & 100 & 29.65 & 34.4 \\
\midrule
H$_2$O & 16 & 42.35 & 44.36 & 44.46 & 69.5 & 43.68 & 25.53 & 91.83 & 23.15 & 6.49 & 52.93 & 24.8 & 43.63 & 68 & 22.78 & 31.79 \\
\midrule
H$_2$O + AQUA-KV & 2.09 & 42.08 & 43.87 & 43.32 & 69 & 44.27 & 25.31 & 92.38 & 23.45 & 6.14 & 53.2 & 24.86 & 41.8 & 68.5 & 22.7 & 30.31 \\
H$_2$O + HIGGS & 2.02 & 41.61 & 41.69 & 44.87 & 69.5 & 43.55 & 25.33 & 91.52 & 22.47 & 6.55 & 52.43 & 24.08 & 41.11 & 68 & 22.24 & 29.13 \\
\midrule
H$_2$O + AQUA-KV & 3.05 & 42.37 & 44.1 & 44.65 & 69.5 & 44.12 & 25.38 & 92.62 & 23.47 & 6.5 & 53.57 & 24.73 & 43.3 & 68 & 22.13 & 31.14 \\
H$_2$O + HIGGS & 3.02 & 42.23 & 43.84 & 45.28 & 69.5 & 43.94 & 25.34 & 91.85 & 23.07 & 6.49 & 53.16 & 24.85 & 42.01 & 68 & 22.41 & 31.45 \\
\midrule
H$_2$O + AQUA-KV & 4.02 & 42.37 & 44.3 & 44.34 & 69.5 & 43.73 & 25.78 & 91.72 & 23.33 & 6.71 & 53.08 & 24.7 & 43.56 & 68 & 22.82 & 31.64 \\
H$_2$O + HIGGS & 4.02 & 42.17 & 44.38 & 43.65 & 69.5 & 43.54 & 25.81 & 91.93 & 22.84 & 6.76 & 53.25 & 24.54 & 42.53 & 68 & 22.13 & 31.58 \\
\end{tabular}
}
\end{table*}

\end{document}